%% file: main.tex
\documentclass{isprs} % isprs class modified 23-04-2019 (Dennis Wittich)

\usepackage{subfigure}
\usepackage{setspace}
\usepackage{geometry} % added 27-02-2014 Markus Englich
\usepackage{epstopdf}
\usepackage[labelsep=period]{caption}  % added 14-04-2016 Markus Englich - Recommendation by Sebastian Brocks
\usepackage[british]{babel} 
\usepackage[hang]{footmisc}

\usepackage{amsmath,amsfonts}
\usepackage{algorithm}
\usepackage{array}

\usepackage{textcomp}
\usepackage{stfloats}
\usepackage{url}
\usepackage{verbatim}
\usepackage{graphicx}
\usepackage{cite}

\usepackage{multirow}
\usepackage{bbding}
\usepackage{times}
\usepackage{epsfig}
\usepackage{amsmath}
\usepackage{amssymb}
\usepackage{caption}
\usepackage{booktabs} 
\usepackage{dsfont}
\usepackage{xcolor}

\usepackage[backref]{hyperref}

\begin{document}

\title{VECTORLLM: HUMAN-LIKE EXTRACTION OF STRUCTURED BUILDING CONTOURS VIA MULTIMODAL LLMS
}
% Leave empty (\version{}) when submitting the full paper
\version{July 2025}

% KAO: Remove extra spacing
% Anonymous submissions, authors' names should not be visible
\author{Tao~Zhang$^a$, Shiqing~Wei$^b$, Shihao~Chen$^a$, Wenling~Yu$^c$, Muying~Luo$^a$, and~Shunping~Ji$^a$$^{\dag}$}

% KAO: Remove extra newline
% Anonymous submissions, authors' affiliations should not be visible
\address{
 $^a$ School of Remote Sensing and Information Engineering, Wuhan University, 129 Luoyu Road, Wuhan 430079, China. \\
 $^b$  College of Oceanography and Space Informatics, China University of Petroleum (East China), Qingdao 266580, China. \\
$^c$ School of Surveying and Geoinformation Engineering, East China University of Technology, Nanchang, 330013, China. \\
$^{\dag}$ Corresponding author \\
E-Mail addresses: zhang\_tao@whu.edu.cn (T. Zhang), wei\_sq@upc.edu.cn (S. Wei), chenshihao@whu.edu.cn (S. Chen),\\ yuwenling@ecut.edu.cn (W. Yu), luomuying@whu.edu.cn (M. Luo), jishunping@whu.edu.cn (S. Ji)
}

% If the corresponding author is NOT the final author, always add a % space before the subsequent comma, i.e.
% first author name\textsuperscript{a,}\thanks{Corresponding author} , % second author name \textsuperscript{b}, etc.
% thanks to Niclas Borlin 05-05-2016

% \commission{XX, }{YY} %This field is optional. If filled, XX and YY should be replaced by adequate numbers. See https://www2.isprs.org/commissions/
% \workinggroup{XX/YY} %This field is optional.
\icwg{}   %This field is optional.

% KAO: Use times symbol
\abstract{
Automatically extracting vectorized building contours from remote sensing imagery is crucial for urban planning, population estimation, and disaster assessment. Current state-of-the-art methods rely on complex multi-stage pipelines involving pixel segmentation, vectorization, and polygon refinement, which limits their scalability and real-world applicability. Inspired by the remarkable reasoning capabilities of Large Language Models (LLMs), we introduce VectorLLM, the first Multi-modal Large Language Model (MLLM) designed for regular building contour extraction from remote sensing images. Unlike existing approaches, VectorLLM performs corner-point by corner-point regression of building contours directly, mimicking human annotators' labeling process. Our architecture consists of a vision foundation backbone, an MLP connector, and an LLM, enhanced with learnable position embeddings to improve spatial understanding capability. Through comprehensive exploration of training strategies including pretraining, supervised fine-tuning, and preference optimization across WHU, WHU-Mix, and CrowdAI datasets, VectorLLM significantly outperformed the previous SOTA methods by 5.6 AP, 7.1 AP, 13.6 AP, respectively in the three datasets. Remarkably, VectorLLM exhibits strong zero-shot performance on unseen objects including aircraft, water bodies, and oil tanks, highlighting its potential for unified modeling of diverse remote sensing object contour extraction tasks. Overall, this work establishes a new paradigm for vector extraction in remote sensing, leveraging the topological reasoning capabilities of LLMs to achieve both high accuracy and exceptional generalization. All the codes and weights will be published for promoting community development. 
}


\maketitle

%\saythanks % added 28-02-2014 Markus Englich

\input{latex/intro}

\input{latex/related_works}

\input{latex/method}

\input{latex/exp}

\input{latex/conclusion}

% {
% 	\begin{spacing}{1.17}
% 		\normalsize
% 		\bibliography{ref} % Include your own bibliography (*.bib), style is given in isprs.cls
% 	\end{spacing}
% }

% {
% 	\bibliographystyle{IEEEtran}
% 	\bibliography{IEEEabrv, main}
% }

{
	\begin{spacing}{1.17}
		\normalsize
		\bibliography{main} % Include your own bibliography (*.bib), style is given in isprs.cls
	\end{spacing}
}

\end{document}

%% file: latex/intro.tex
\section{Introduction}
\label{sec:intro}

Automatically and accurately extracting buildings from remote sensing and satellite image is an important task that serves as a crucial component for many downstream applications, including urban planning, population estimation, disaster assessment, map updating, and others. Over the past decade, with the development of deep learning, fully automated building extraction methods~\cite{mafcn, buildmapper, clpcnn, zhang2024p2pformer, wei2024lines, polygonrnn, polygonrnn++, polyworld, polymapper, ffl, xu2023hisup, wang2023image, xia2024vectorizing, luo2024sam} have achieved tremendous progress. A large number of deep learning-based architectures~\cite{mafcn, buildmapper, clpcnn, zhang2024p2pformer, wei2024lines, polygonrnn, polygonrnn++, polyworld, polymapper, ffl, xu2023hisup} have been proposed to automatically extract vectorized building contours from aerial and satellite imagery.

\sloppy{However, current state-of-the-art methods~\cite{mafcn, zhang2024p2pformer, wei2024lines, xu2023hisup}, whether based on mask segmentation or polygon regression approaches, all involve complex pipelines and multiple proxy tasks. For example,~\cite{mafcn, xu2023hisup} include multiple stages such as pixel segmentation, vectorization, and polygon refinement, while~\cite{buildmapper, zhang2024p2pformer, wei2024lines} involve complex proxy tasks including vertex estimation, redundant point removal, and topological relationship reconstruction. The complexity and unique designs of these methods greatly limit further scaling of models and data.}

Recently, Large Language Models (LLMs)~\cite{bai2023qwen, yang2025qwen3, cai2024internlm2, vicuna2023} have achieved remarkable progress, demonstrating intelligence comparable to professional humans in various domains such as coding and mathematics. The intelligence and logical reasoning capabilities they exhibit are groundbreaking and unprecedented in previous vision models~\cite{dosovitskiy2020image, sam}. Multi-modal Large Language Models (MLLMs)~\cite{chen2024internvl, chen2024expanding, chen2024far, Qwen2VL, bai2023qwenvl} efficiently construct general-purpose multi-modal intelligent agents by aligning features from other modalities to the language space, similarly demonstrating amazing logical reasoning capabilities and task emergence abilities on complex visual tasks.

\textbf{\textit{A native question arises: can we build an MLLM for vectorized building contour extraction task that can output regular vector-form building corner-point by corner-point like human annotators do?}} When human annotators label building contours,  building recognition capabilities as well as topological reasoning abilities are required, and these are precisely the capabilities that LLMs and MLLMs excel at.

In this paper, we introduce LLMs into the vector extraction field and use regular vector contour extraction of buildings as a validation task. As pioneers in exploring LLMs for vector contour extraction from remote sensing images, we focus on investigating whether the powerful topological reasoning and generalization capabilities of LLMs can bring new prospects to the vector extraction field in remote sensing. We simplify the task of regular building contour extraction that MLLMs handle: given building bounding boxes, we perform point-by-point regression of the regular vector contours of buildings. This simplification brings two benefits: 1) it excludes the challenges that dense perception tasks pose to MLLMs, which is one of the challenges that the MLLM field currently focuses on addressing~\cite{Qwen2VL, zhu2025internvl3}, and 2) it excludes the additional challenges of collaborative training brought by differences in image sizes and resolutions across different datasets~\cite{crowdai, ji2018fully, buildmapper}, thereby allowing us to deeply explore the benefits brought by scaling data. 
%We leave the resolution of these challenges and the unified modeling of more complex vector extraction tasks to our future works.

\begin{figure*}
  \centering
  \includegraphics[width=1.00\linewidth]{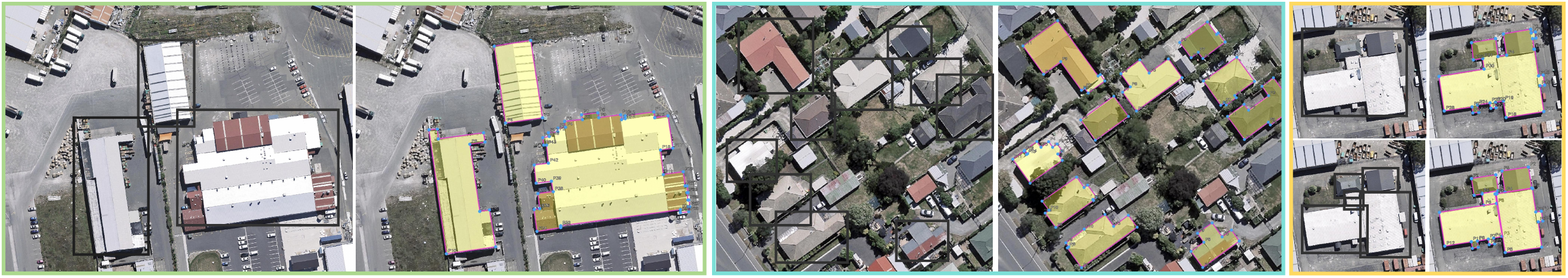}
  \includegraphics[width=1.00\linewidth]{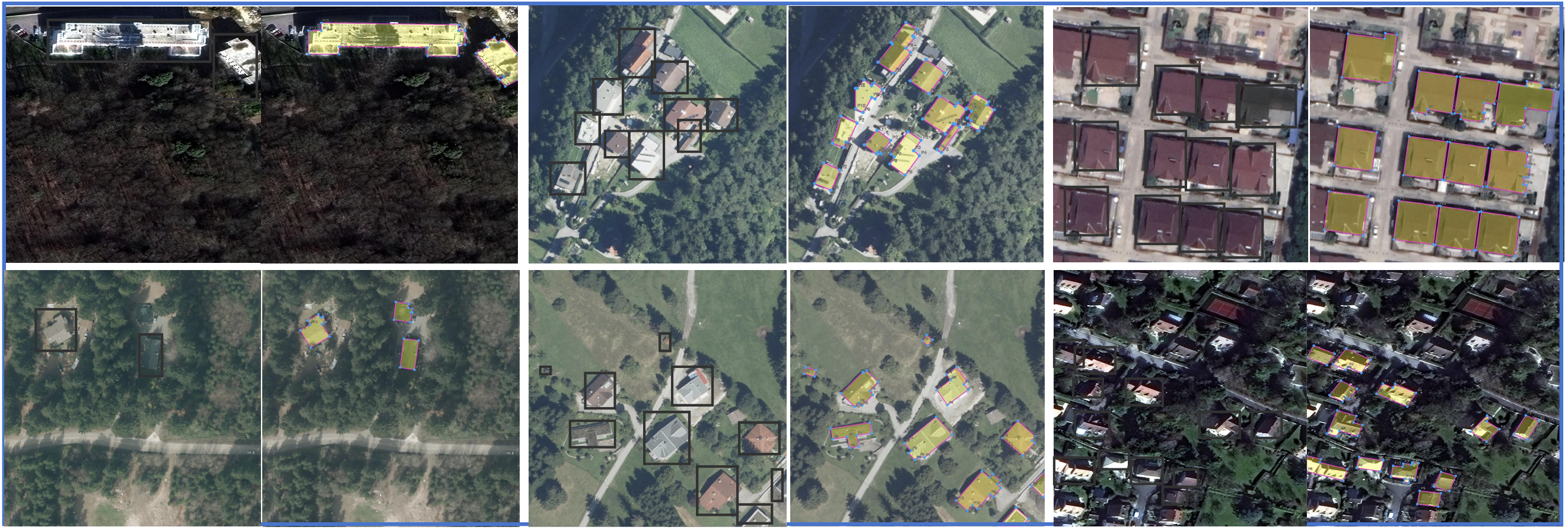}
  \includegraphics[width=1.00\linewidth]{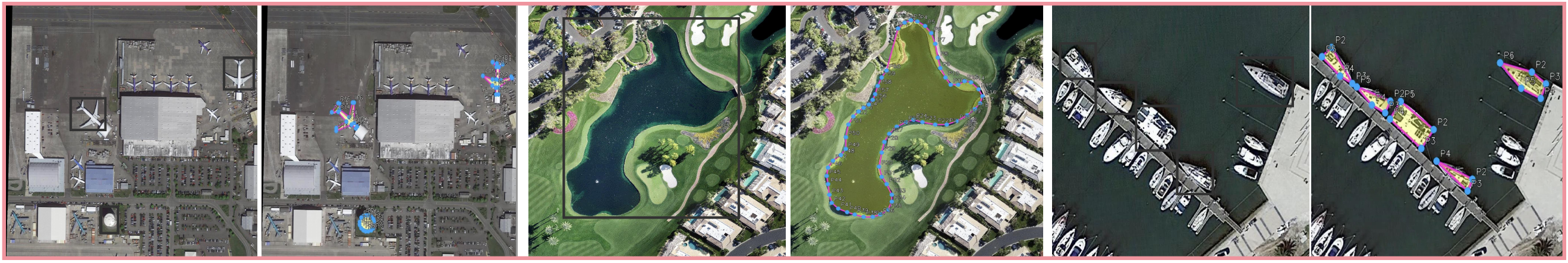}
  \caption{\textbf{The strong capability of VectorLLM.} The black bounding boxes are manually provided by users. The subfigures within the green and cyan borders in the first row demonstrate that VectorLLM can accurately extract regular contours of buildings without any post-processing processes. The subfigures within the yellow borders show that VectorLLM can flexibly handle multi-granularity vector building contour extraction tasks, such as extracting regular contours of building clusters or individual buildings. The subfigures within the blue borders in the second row demonstrate that VectorLLM works well on diverse remote sensing images, using the same weights without requiring targeted training on specific datasets. The third row showcases VectorLLM's strong generalization performance, exhibiting remarkable zero-shot performance on unseen objects such as airplanes, circular oil tanks, water bodies, and ships.}
  \label{fig:teaser}
\end{figure*}

\begin{figure*}
  \centering
  \includegraphics[width=0.95\linewidth]{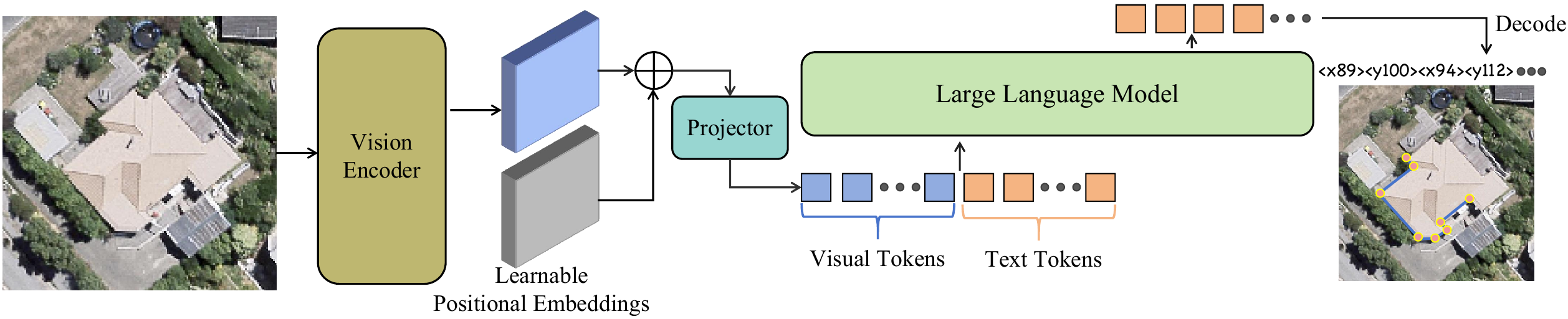}
  \caption{\textbf{The architecture of VectorLLM.} VectorLLM has a streamlined architecture, consisting of a vision encoder, learnable positional embedding, MLP projector, and LLM. VectorLLM will generate regular building contours point by point in an auto-regressive manner.}
  \label{fig:architecture}
\end{figure*}

We propose VectorLLM, the first MLLM for regular building contour extraction, composed of a vision foundation backbone, an MLP connector, and an LLM. To enhance VectorLLM's positional modeling capability, we introduce learnable position embeddings to inject positional information into the vision tokens output by the vision foundation backbone. Notably, we preserve the simplicity of the architecture as much as possible without introducing any task-specific designs, so that VectorLLM can easily be extended and unified for modeling on more vector extraction tasks in the future, such as road extraction and roof fine-structure extraction.

In the field of large models, data and training strategies are far more important than unique architectural designs. Therefore, this paper focuses on exploring reasonable data organization methods and training approaches to unleash the amazing potential of VectorLLM in the vector extraction field. We explore the complete training pipeline of MLLM on vector extraction tasks, including pretraining, supervised fine-tuning, and preference optimization. At each training stage, we thoroughly investigate the effects of different data organization formats and training strategies.

Through collaborative training on WHU~\cite{ji2018fully}, WHU-Mix~\cite{buildmapper}, and CrowdAI~\cite{crowdai}, VectorLLM demonstrates strong potential for vector extraction tasks. VectorLLM achieves 79.4 AP on the WHU dataset, significantly outperforming current SOTA methods P2PFormer~\cite{zhang2024p2pformer} and Line2Poly~\cite{wei2024lines}. Furthermore, VectorLLM achieves 68.4 AP and 56.4 AP on the WHU-Mix test-1 set (in-domain) and test-2 set (out-of-domain) respectively. VectorLLM achieves 79.6 AP on the CrowdAI dataset, similarly surpassing current SOTA models in performance. It's worth noting that VectorLLM is co-trained on three building datasets and tested using the same weights, without separate fine-tuning on any specific dataset. Finally, we are pleasantly surprised to find that VectorLLM shows good zero-shot results on unseen targets such as water bodies, aircraft, and other remote sensing objects (see Fig.~\ref{fig:teaser}).

To summarize, our paper makes three contributions:
\begin{itemize}
    \item We introduce LLMs to remote sensing vector extraction tasks and construct VectorLLM, the first multi-modal large model for remote sensing vector extraction.

    \item We thoroughly explore training methods, including the effects of pretraining, SFT, and preference optimization on VectorLLM, as well as the impact of data organization methods at each training stage.

    \item VectorLLM achieves SOTA performance on WHU, WHU-Mix  and CrowdAI datasets and demonstrates remarkable generalization capabilities, providing a solid baseline for future unified modeling of remote sensing vector extraction tasks.
\end{itemize}

\begin{figure*}
  \centering
  \includegraphics[width=1.0\linewidth]{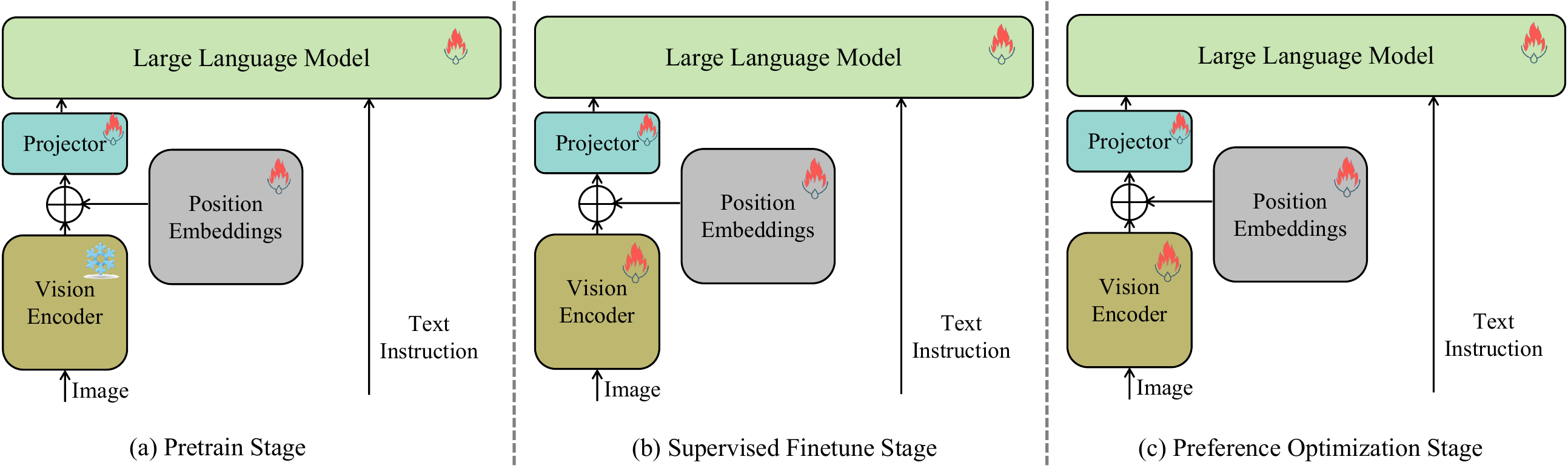}
  \caption{\textbf{Training stages of VectorLLM.} The training of VectorLLM consists of pretraining, supervised finetuning, and preference optimization stages.}
  \label{fig:train}
\end{figure*}

%% file: latex/related_works.tex
\section{Related Works}
\label{sec:related_works}

\subsection{Building Extraction Methods}

There are a lot of methods~\cite{r3, r4, r5, r6, r7, r8, r9, mafcn, r16, r17, r18, polytransform, r20, ffl, r23, r24, r25, r26, r27, r28, clpcnn, r34, e2ec, buildmapper, zhang2024p2pformer} focusing on extracting buildings from remote sensing images, which is challenging and important for downstream applications. \cite{r3, r4, r5, r6, r7, r8, r9, mafcn, r16, r17, r18, polytransform, r20, ffl} have designed various mask segmentation networks for building extraction. However, they suffer from problems such as imprecise contour segmentation and fragmented internal segmentation results. These challenges make it difficult for these methods to be directly applied to downstream tasks such as mapping~\cite{xia2024vectorizing, xia2024video} and urban planning, which require precise regular vector-form contours. Even SAM~\cite{kirillov2023segment}, trained on massive datasets and its derivatives~\cite{luo2024sam, chen2024rsprompter} in remote sensing, cannot adequately meet the requirements of downstream tasks. 

Some works~\cite{e2ec, buildmapper, zhang2024p2pformer, polyworld} obtain building contours in vector form by directly regressing edge point coordinates without relying on post-processing procedures. However, these works differ significantly from human annotators' annotation processes, thus suffering from redundant points and unreasonable topological errors. Polygon-RNN~\cite{polygonrnn} and Polygon-RNN++~\cite{polygonrnn++} use RNN to achieve point-by-point prediction, similar to human annotators. However, due to the limitations of their outdated architectures, they are affected by error accumulation and thus cannot achieve satisfactory results. In this paper, we introduce the Large Language Model (LLM), which shows powerful reasoning ability, into the building contour extraction task. Based on a pre-trained LLM, we build a framework that can extract building contours like human annotators, demonstrating strong topological stability and generalization capabilities.

\subsection{Multimodal Large Language Models}

Although large language models~\cite{floridi2020gpt, touvron2023llama, dubey2024llama, team2023internlm, cai2024internlm2, bai2023qwen, yang2024qwen2} have demonstrated powerful reasoning and generalization capabilities, they cannot understand visual signals like images, which limits their application in downstream tasks. Many works~\cite{liu2023llava, liu2023improvedllava, chen2024expanding, chen2024far, chen2024internvl, bai2023qwenvl, Qwen2VL} have attempted to endow LLMs with the ability to understand visual signals. Among these, LLaVA~\cite{liu2023llava, liu2023improvedllava}, with concise architectural design and powerful performance, has inspired and led many subsequent works. LLaVA employs a simple MLP layer to project image features obtained from a pre-trained CLIP~\cite{radford2021learning} backbone into the language feature space, enabling large language models to understand image inputs. LLaVA acquired strong image and text question-answering capabilities through pretraining and supervised fine-tuning stages. Current state-of-the-art multimodal large language models (MLLMs) like InternVL~\cite{chen2024internvl, chen2024expanding, chen2024far} and Qwen-VL~\cite{Qwen2VL, bai2023qwenvl} series have achieved outstanding image understanding and dialogue capabilities by introducing more high-quality data, improving training strategies, image processing approaches, and visual positioning encoding techniques.

Some MLLMs~\cite{lai2024lisa, ren2024pixellm, rasheed2024glamm, zhang2024omg, yuan2025sa2va, zhang2025pixel} achieve text-driven pixel segmentation by combining multimodal large language models (MLLMs) with a segmentation expert. \cite{yuan2024osprey, lin2024drawandunderstand, rasheed2024glamm, zhang2024omg} grant MLLMs region-level understanding and dialogue capabilities by designing different region encoding modules. However, no work has yet attempted to use LLMs for vector extraction tasks in remote sensing. In this paper, we comprehensively explore how to construct a multi-modal large language model for vector extraction tasks in remote sensing, using building extraction as an example.

\subsection{Remote Sensing Multimodal Large Language Models} 

Due to the lack of corresponding training data and domain knowledge, general MLLMs such as LLaVA~\cite{liu2023llava, liu2023improvedllava}, InternVL2~\cite{chen2024expanding, chen2024far, chen2024internvl}, and Qwen2-VL~\cite{Qwen2VL} perform poorly in the remote sensing domain. Many works~\cite{wang2024ringmogpt, pang2024h2rsvlm, muhtar2024lhrs, kuckreja2024geochat, zhang2024earthgpt, muhtar2024lhrs, zhan2024skyeyegpt, luo2024skysensegpt, li2024vrsbench, zhang2024rs5m, wang2024skyscript} have attempted to establish a remote sensing MLLM to handle various tasks such as question answering, change detection, segmentation, and detection. \cite{guo2024remote, liu2024change, singh2024geollm} construct MLLMs capable of handling remote sensing tasks by combining powerful general MLLMs like GPT-4 with various remote sensing specialist models. \cite{muhtar2024lhrs, zhan2024skyeyegpt, luo2024skysensegpt, li2024vrsbench, zhang2024rs5m, wang2024skyscript} collect large-scale remote sensing instruction tuning datasets and construct MLLMs that perform well in the remote sensing domain. Meanwhile, \cite{luo2024skysensegpt, wang2024ringmogpt, zhou2024geoground, zhang2024popeye} propose remote sensing MLLMs capable of text-driven detection or mask segmentation. \cite{elgendy2024geollava, zhang2024popeye, irvin2024teochat} introduced MLLMs into downstream tasks, such as change detection with text description. However, current remote sensing MLLMs can only achieve mask segmentation or coarse-grained bounding box detection, and there is currently no remote sensing MLLM capable of fine-grained vector extraction. In this paper, we construct the first remote sensing MLLM with fine-grained vector extraction capability and explore in detail the MLLM's pretraining, supervised finetuning (SFT), and preference optimization (PO) stages.

%% file: latex/method.tex
\section{Methodology}
\label{sec:method}

\begin{figure*}
  \centering
  \includegraphics[width=0.98\linewidth]{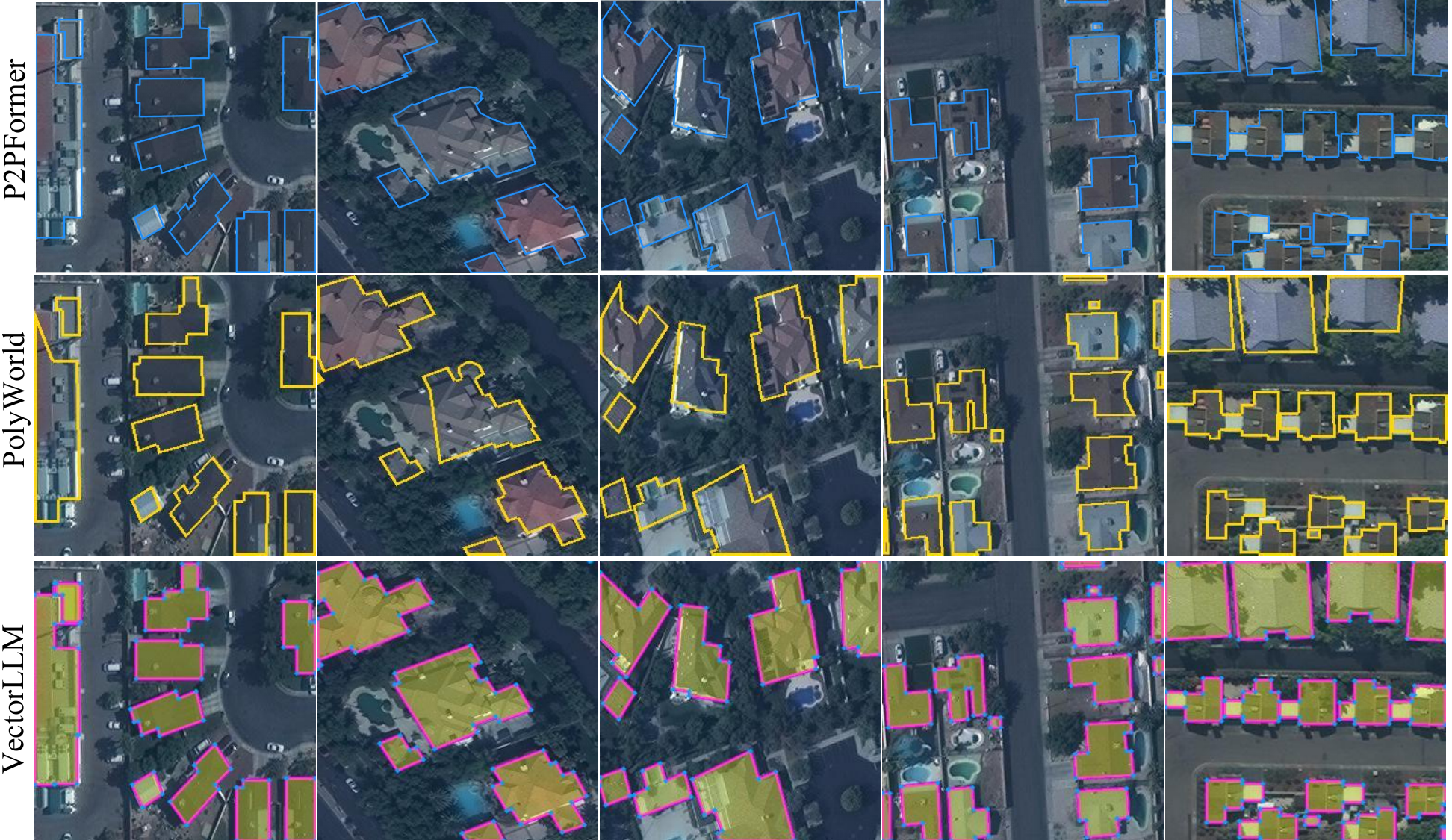}
  \caption{\textbf{The visualization results on CrowdAI dataset.} Although the low image resolution and low brightness of the CrowdAI dataset limit VectorLLM's performance, VectorLLM still demonstrates more accurate and stable topology prediction results compared to PolyWorld and P2PFormer.}
  \label{fig:comparision_visualize_crowdai}
\end{figure*}

In this section, we introduce VectorLLM's architecture and training strategy in detail. In section~\ref{sub_sec:network}, we provide a detailed introduction to VectorLLM's network architecture. In sections~\ref{sub_sec:pretrain}, \ref{sub_sec:sft} and \ref{sub_sec:po}, we explain VectorLLM's training strategy and data organization, which are even more important than the network architecture.

\subsection{Network Architecture}
\label{sub_sec:network}

As shown in Fig.~\ref{fig:architecture}, our proposed VectorLLM consists of a visual encoder, learnable position embeddings, a projector, and a large language model. Images are first resized to a fixed size and processed through a pre-trained visual encoder to obtain vision features. Then, position information is injected into the vision features by adding the learnable position embeddings. The projector maps the vision features into the language feature space to obtain vision tokens, which serve as input to the large language model for providing image information. Finally, the large language model outputs corresponding coordinates in text form based on the vision tokens and text instruction.

\noindent\textbf{Vision Encoder.} We use the pre-trained RADIO~\cite{ranzinger2024radio} VIT~\cite{dosovitskiy2020vit} backbone as the vision encoder, which is distilled from CLIP~\cite{radford2021learning}, SAM~\cite{sam} and DINOv2~\cite{oquab2023dinov2}. The input image is resized to 128$\times$128 and produces image features downsampled by a factor of 16 through the vision encoder.

\noindent\textbf{Position Embeddings.} We employ learnable embeddings to embed positional information into the image features. The learnable position embeddings can help the LLM understand each pixel's semantic and positional information.

\noindent\textbf{Projector.} Similar to ~\cite{liu2023improvedllava, chen2024far}, the projector is a simple MLP layer. The projector maps vision features into language space, enabling the large language model to understand image information.

\noindent\textbf{Large Language Model.} The large language model is pre-trained on massive text data, possessing strong logical reasoning and instruction-following capabilities. To enable the large language model to output coordinates, we added special tokens to the large language model's vocabulary (from [x0] to [x(W-1)] and from [y0] to [y(H-1)]), totaling H+W tokens. The large language model accepts vision tokens and text tokens as input and outputs text format responses, which use these special tokens to represent geometric coordinates.

\subsection{Pretraining}
\label{sub_sec:pretrain}

\noindent\textbf{Pretraining Tasks.} Pretraining is widely used in natural language processing and computer vision to inject fundamental knowledge into models. In LLaVA~\cite{liu2023llava}, the image caption task is used as the pretraining task to learn the projector that aligns CLIP image features to language space, thus enabling LLM to understand images. 

However, unlike LLaVA which only aligns vision features to text space, in the pretraining stage we not only need to enable the large language model to understand remote sensing images, but also need to inject geometric reasoning capabilities into it. The pretrained vision encoder (such as RADIO~\cite{ranzinger2024radio} or DinoV2~\cite{oquab2023dinov2}) already possesses strong modeling capabilities for natural images and remote sensing images. Therefore, during the pretraining stage, we freeze the vision encoder, training the MLP projector, the learnable positional embeddings, and the large language model.

We directly concatenate images with their corresponding building contour coordinates as pretraining data, and compute the next token prediction loss only on text tokens. A sample of pretraining data is as follows:

``\textit{Input: [image]\textbackslash n[x85][y32][x160][y63][x135][y122][x176]
[y139][x154][y191][x103][y169][x111][y150][x46][y124]$\\$
[x85][y32]}."

It is worth noting that we found randomly shuffling the starting point of contours helps improve pretraining effectiveness, as this prevents the topological reasoning of subsequent points from overly relying on a fixed starting point.

\noindent\textbf{Training Strategy.} As shown in Fig.~\ref{fig:train}, during the pretraining stage, we freeze the vision expert and train the projector, the learnable positional embeddings, and the large language model. We use the standard next-token-prediction loss to maximize the probability of the model predicting answers based on the image and the instruction. Specifically, the coordinates of all previous corner points in a pretraining sample are used to predict the next corner point, starting from the coordinate prediction of the first corner point.

\subsection{Supervised Finetune}
\label{sub_sec:sft}

\noindent\textbf{Supervised Finetune Task.} In this stage, the model will learn to output corresponding answers according to user instructions. We reorganize the building extraction datasets into question answering format to serve as instruction tuning data for the SFT stage. We use the building's top-left corner point (the point closest to the image's top-left corner) as the starting point, arrange the building's corner points in a clockwise direction, and finally return to the starting point to generate the answer. The instruction data is organized in the following format:

``\textit{Input: [image]\textbackslash nPlease extract the regular vector contour of the central building in the image, start from the left top corner and in clockwise.}$\\$
\textit{Output: [x85][y32][x160][y63][x135][y122][x176][y139]$\\$
[x154][y191][x103][y169][x111][y150][x46][y124][x85]$\\$
[y32]}."

Unlike pretraining data, we found that selecting building starting points according to certain fixed rules in SFT data brings slight performance improvements and stability in model predictions.

\noindent\textbf{Train Strategy.} As shown in Fig.~\ref{fig:train}, in addition to the projector and learnable positional embeddings, we also fine-tune both the vision expert and the large language model. We similarly use the standard next token prediction loss to train the model.

\begin{figure*}
  \centering
  \includegraphics[width=0.98\linewidth]{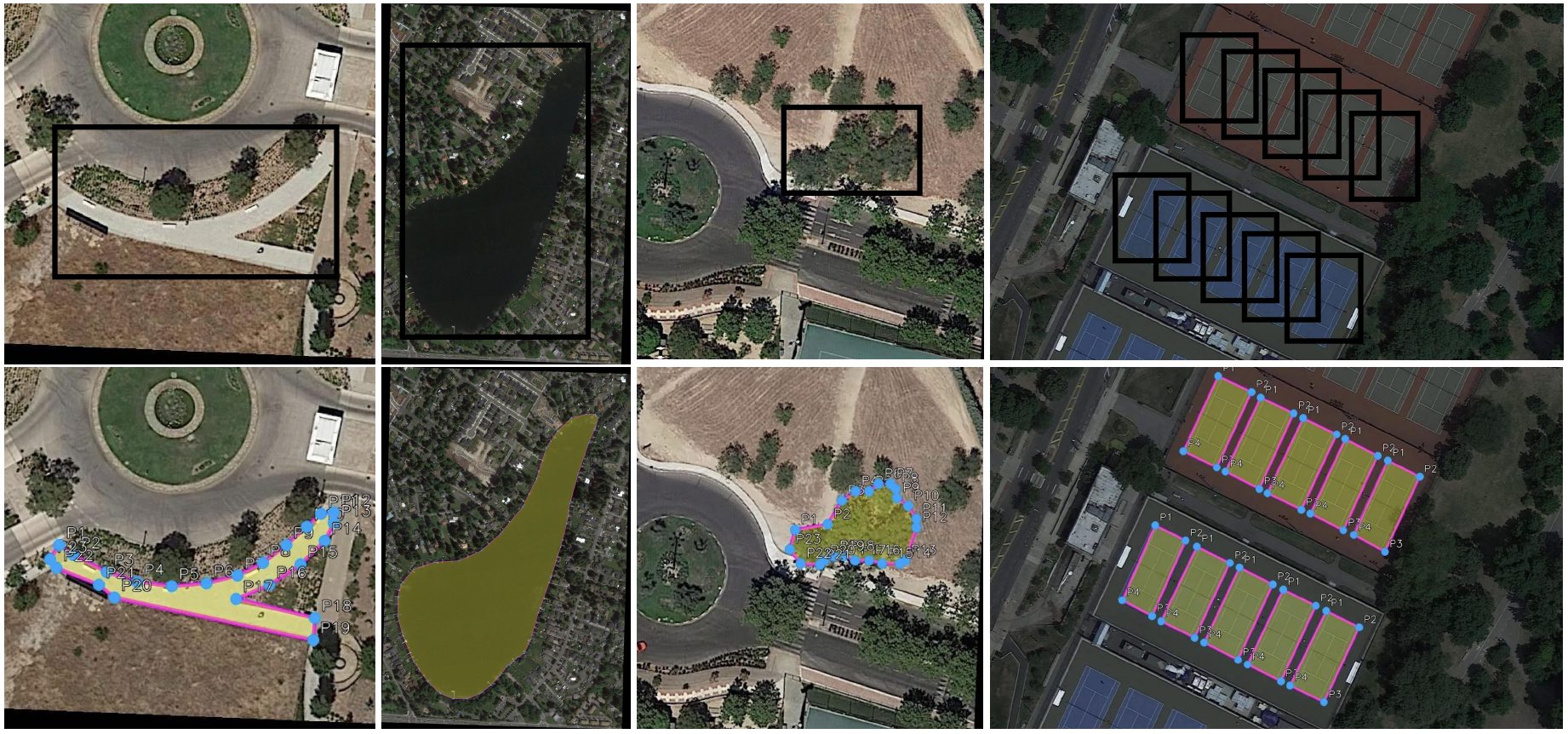}
  \caption{\textbf{The visualization results of VectorLLM's zero-shot prediction results.} VectorLLM demonstrates strong zero-shot performance on feature types that do not exist in the training set, such as roads, water bodies, vegetation, and playgrounds. The black bounding boxes in the first row are manually provided by users.}
  \label{fig:other_rs_visualize}
\end{figure*}

\subsection{Preference Optimization}
\label{sub_sec:po}

After supervised tuning, the model can accurately and robustly extract regular building contours from images automatically. However, there are still some minor problems, such as occasional repeated outputs, missing small lines, and errors caused by interference from roof planes. To address and optimize these issues and make the model's output more aligned with user preferences, we perform direct preference optimization~\cite{rafailov2024direct} (DPO) on the model.

\noindent\textbf{Data Generation.} The most important aspect of direct preference optimization  is collecting preference data (rejected and chosen answer pairs). We design simple rules to automatically obtain preference data. First, we perform inference on the training set using the SFT model, and use predictions with an IoU less than 0.8 with the labels as rejected answers, with the corresponding labels serving as chosen answers. Furthermore, to expand the preference data with more variety, we randomly corrupt the labels of complex buildings (more than 15 points) and large buildings by deleting points or inserting random points to generate rejected answers, using the uncorrupted labels as chosen answers. The DPO data is organized in the following format:

``\textit{Input: [image]\textbackslash nPlease extract the regular vector contour of the central building in the image, start from the left top corner and in clockwise.}$\\$
\textit{Rejected Answer: [x83][y40][x96][y38][x95][y30][x116]$\\$
[y26][x121][y49][x111][y51][x112][y58][x106][y60][x115]$\\$
[y107][x128][y104][x134][y137][x121][y139][x126][y165]$\\$
[x140][y187][x83][y197][x83][y40]}$\\$
\textit{Chosen Answer: [x85][y47][x97][y45][x96][y37][x115]$\\$
[y34][x120][y55][x111][y57][x112][y64][x107][y65][x115]$\\$
[y107][x126][y105][x131][y135][x120][y137][x125][y160]$\\$
[x133][y159][x137][y180][x85][y190][x85][y47]}."

\noindent\textbf{Train Strategy.} We use the standard DPO loss to train the model:
\begin{equation}
    \mathcal{L}_{dpo} = -log~\sigma(\beta~log\frac{\pi_{\theta}(y_{c}~|~x)}{\pi_{0}(y_{c}~|~x)} - \beta~log\frac{\pi_{\theta}(y_{r}~|~x)}{\pi_{0}(y_{r}~|~x)})~,
\end{equation}
where $\beta$ is the KL penalty coefficient, and $x$, $y_{c}$, and $y_{r}$ are user instruction, chosen answer, and rejected answer, respectively. The policy model $\pi_{\theta}$ is initialized from the model $\pi_{0}$. We freeze the model obtained from the supervised fine-tuning stage as $\pi_{0}$. Our objective is to optimize the model $\pi_{\theta}$ using the DPO loss.

\subsection{Bounding Box Access}

VectorLLM is designed to delineate the contour of a prominent building in the image, because current LLMs show incompetence in directly recognizing fine-grained multiple object instances up to 100 in one image, largely due to the limited tokens. Therefore, we use a bounding box detector to isolate each building instance. In this paper, FCOS~\cite{fcos} with Swin-L~\cite{swin} backbone is used, the detected bounding boxes are slightly enlarged to contain necessary background information. During training, we crop each building instance from the original training images as input to VectorLLM, while during testing, all building instances in the test images are pre-cropped using FCOS.

\begin{figure}[ht]
  \centering
  \includegraphics[width=1.0\linewidth]{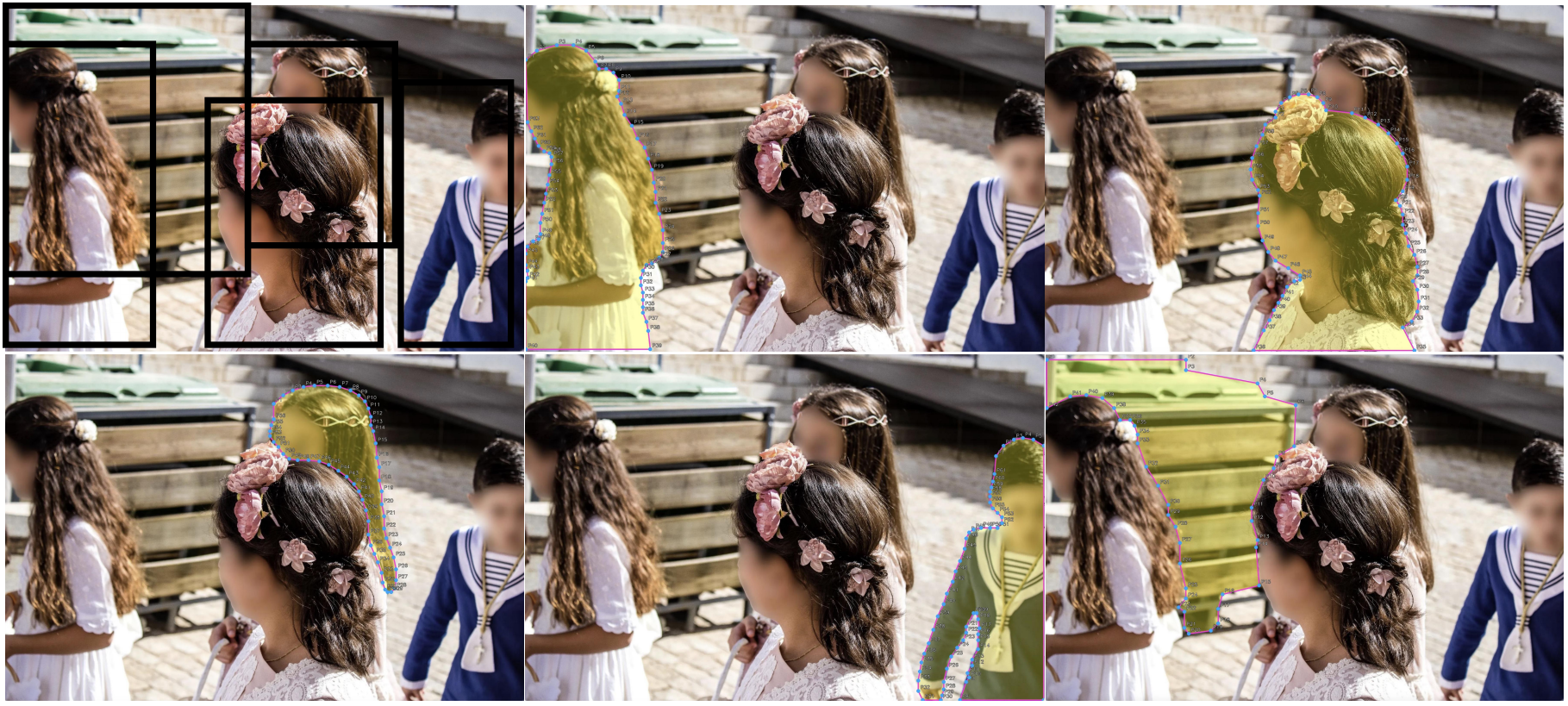}
  \includegraphics[width=1.0\linewidth]{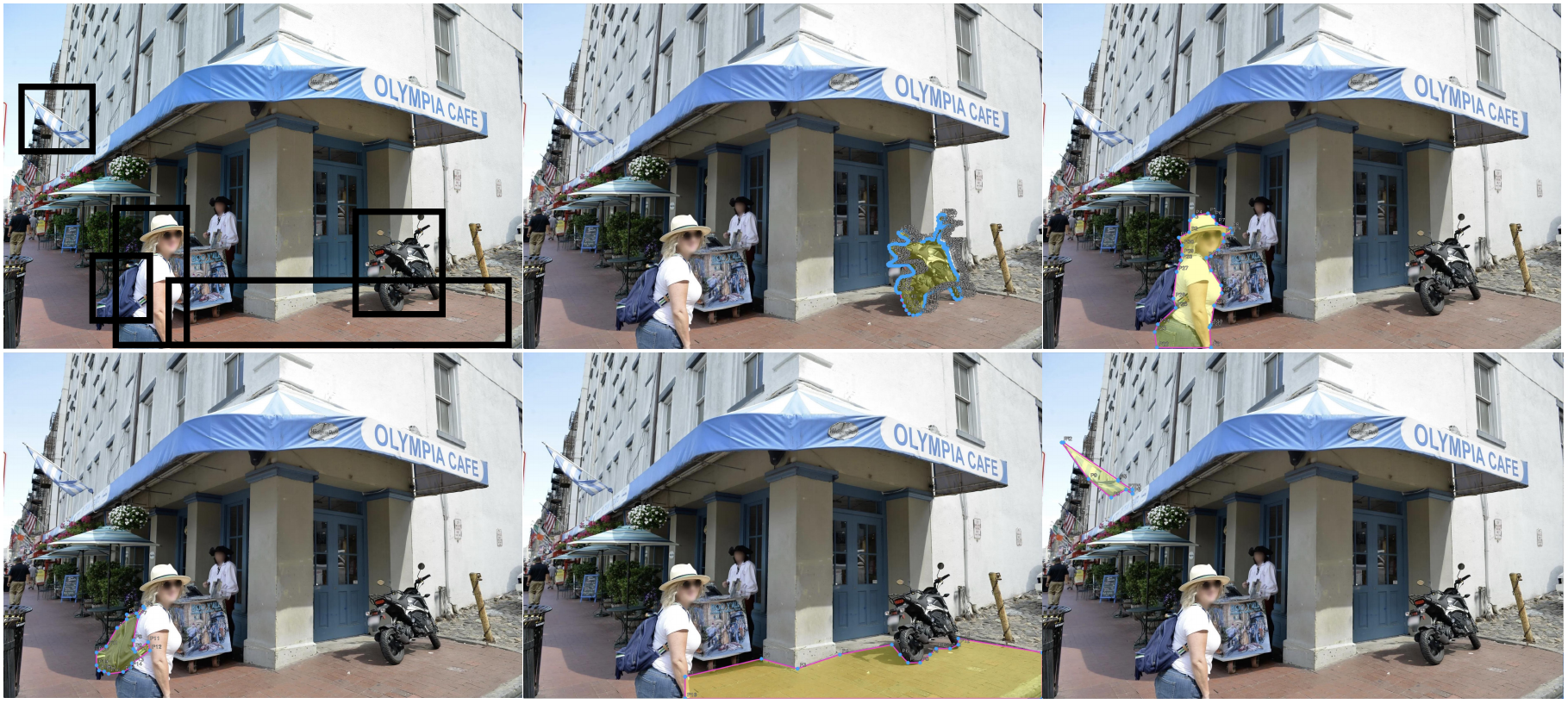}
  \includegraphics[width=1.0\linewidth]{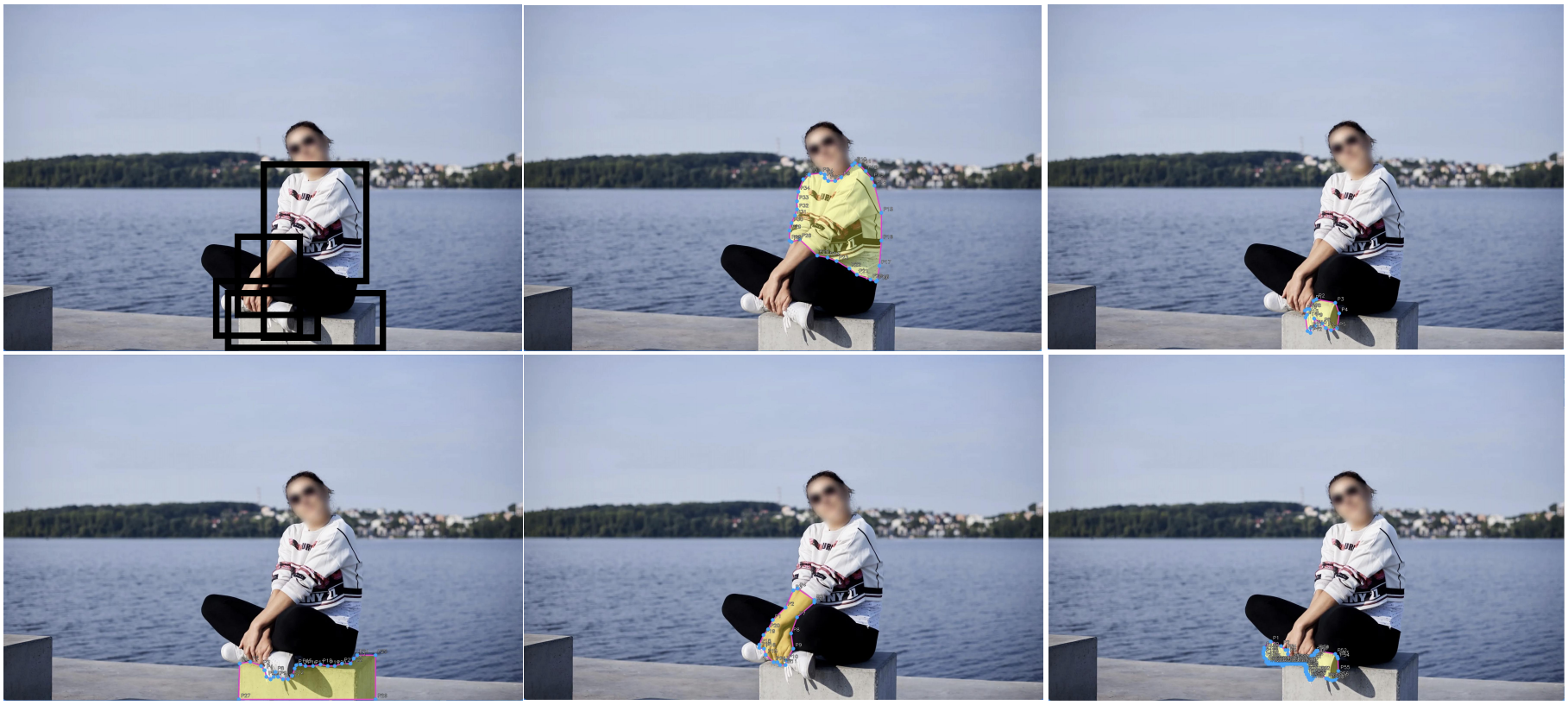}
  \caption{\textbf{The prediction results of VectorLLM on general objects.} All images come from SAM~\cite{sam}. VectorLLM possesses strong vector contour extraction capabilities for multi-granularity general objects. Each image following the image with multiple bounding boxes shows the vector-format contour of a boxed instance. Zoom in to view details.}
  \label{fig:general_obj_visualize}
\end{figure}

%% file: latex/exp.tex
\begin{table}[t]
 \centering
 \caption{\textcolor{black}{Results on the test set of WHU dataset.}}
 \resizebox{0.5\textwidth}{!}{
 \begin{tabular}{l|cccc}
    \toprule[0.2em]
     Method & $mAP$ & $AP_{50}$ & $AP_{75}$ & $AR$ \\
	\hline
	Mask-RCNN~\cite{maskrcnn} & 65.3 & 90.0 & 77.1 & 70.7\\
	QueryInst~\cite{queryinst} & 62.6& 78.6 & 71.1 & 67.1\\
	YOLACT~\cite{yolact} & 58.5 & 76.5 & 69.8 & 64.2\\
	SOLO~\cite{solo} & 67.9 & 87.5 & 79.2 & 71.6\\
	E2EC~\cite{e2ec} & 71.3 & 90.4 & 81.4 & 75.1\\
        CLP-CNN~\cite{clpcnn} & 72.6 & 90.9 & 82.6 & 78.0 \\ 
        BuildMapper~\cite{buildmapper} & 73.6 & 89.0 & 81.6 & 78.9 \\
	P2PFormer~\cite{zhang2024p2pformer} &  72.7 & 91.1 & 83.2 & 77.8 \\
        Line2Poly~\cite{wei2024lines} & 73.8 & 89.4 & 82.1 & 79.7 \\ 
    \hline 
    VectorLLM-0.6B & 78.3 & 91.7 & 86.0 & 82.8 \\
    VectorLLM-1.7B & 79.4 & 91.6 & 85.9 & 83.8\\
    \hline
    VectorLLM-0.6B~(oracle bbox) & 88.1 & 98.9 & 97.3 & 92.9\\
    VectorLLM-1.7B~(oracle bbox) & 90.2 & 98.9 & 98.3 & 94.3 \\
    \bottomrule[0.1em]
\end{tabular}
}
\label{tab:whu}
\end{table}

\begin{table*}[t]
 \centering
 \caption{\textcolor{black}{Results on the WHU-Mix dataset.}}
 \begin{tabular}{l|cccc|cccc}
    \toprule[0.2em]
     \multirow{2}{*}{Method} & \multicolumn{4}{c|}{Test 1} & \multicolumn{4}{c}{Test 2} \\
     ~ & $mAP$ & $AP_{50}$ & $AP_{75}$ & $AR$ & $mAP$ & $AP_{50}$ & $AP_{75}$ & $AR$ \\
    \hline
    Mask R-CNN~\cite{maskrcnn} &  47.0 & 67.0 & 53.2 & 53.7 & 46.1 & 73.9 & 49.0 & 54.8 \\
    YOLACT~\cite{yolact} & 42.3 & 65.7 & 47.2 & 49.7 & 41.3 & 71.3 & 42.3 & 50.6 \\
    SOLO~\cite{solo} & 57.1 & 83.2 & 65.1 & 63.9 & 45.3 & 74.3 & 47.9 & 54.7 \\
    Mask2Former~\cite{mask2former} & 57.6 & 84.9 & 64.6 & 66.1 & 48.2 & 78.1 & 51.1  & 53.2 \\
    PolarMask~\cite{polarmask} &  44.8 & 69.1 & 50.7 & 52.2 & 39.1 & 66.3 & 40.5 & 45.1 \\
    Deep Snake~\cite{deepsnake} & 55.3 & 82.1 & 63.0  & 61.8 & 46.9 & 73.9 & 51.5  & 54.8 \\
    BuildMapper~\cite{buildmapper} & 58.0 & 82.6 & 65.9 & 65.5 & 48.1 & 73.2 & 52.0 & 56.6 \\
    Line2Poly~\cite{wei2024lines} & 58.6 & 81.9 & 66.2 & 65.5 & 48.9 & 73.3 & 52.8  & 58.4 \\
    P2PFormer~\cite{zhang2024p2pformer}& 61.3 & 87.6 & 69.8 & 70.2 & 53.2 & 80.9 & 58.0 & 60.0 \\
    \hline
    VectorLLM-0.6B & 67.8 & 86.5 & 72.4  & 76.5 & 55.6 & 84.1 & 59.4 & 68.2 \\
    VectorLLM-1.7B & 68.4 & 86.9 & 72.5 & 77.0 & 56.4 & 84.7 & 59.8 & 69.0 \\
    \hline 
    VectorLLM-0.6B~(oracle bbox) & 78.0 & 96.6 & 85.0 & 86.4 & 64.4 & 96.0 & 73.8 & 72.6 \\ 
    VectorLLM-1.7B~(oracle bbox) & 78.5 & 96.7 & 86.0 & 86.5 & 64.6 & 96.6 & 74.2 & 73.0 \\
    \bottomrule[0.1em]
\end{tabular}
\label{tab:whu-mix}
\end{table*}

\begin{table}[t]
 \centering
 \caption{Results on the CrowdAI dataset.}
  \resizebox{0.5\textwidth}{!}{
 \begin{tabular}{l|ccc}
    \toprule[0.2em]
	Method &  $mAP$ & $AP_{50}$ & $AP_{75}$\\ \hline
	Mask R-CNN~\cite{maskrcnn} & 41.9 & 67.5 & 48.8\\
	PANet~\cite{panet} &  50.7 & 73.9 & 62.6\\
	Mask2Former~\cite{mask2former} & 63.0 & 91.5 & 72.9 \\
        HigherNet-DST~\cite{he2024highernet} & 68.5 & 88.4 & 77.5 \\
	\hline
        PolyMapper~\cite{polymapper}  & 55.7 & 86.0 & 65.1\\
	BuildMapper~\cite{buildmapper} &  63.9 & 90.1 & 75.0\\
        FFL~\cite{ffl} & 61.7 & 87.6 & 71.4\\
	PolyWorld~\cite{polyworld} & 63.3 & 88.6 & 70.5\\
    HiT~\cite{zhang2024hit} & 64.6 & 91.9 & 78.7 \\
    P2PFormer~\cite{zhang2024p2pformer} & 66.0 & 91.1 & 77.0\\
    \hline
    VectorLLM-0.6B & 78.0 & 90.0 & 79.6 \\
    VectorLLM-1.7B & 79.6 & 91.5 & 81.3  \\
    \hline
    VectorLLM-0.6B (oracle bbox) & 86.1 & 93.0 & 89.3  \\
    VectorLLM-1.7B (oracle bbox) & 87.7 & 94.2 & 90.4  \\
 \bottomrule[0.1em]
 \end{tabular}
 }
 \label{tab:crowdai}
\end{table}

\begin{table}[t]
 \centering
 \caption{Ablation studies on the learnable positional embedding.}
 \begin{tabular}{c|cccc}
    \toprule[0.2em]
     Positional Embedding & $mAP$ & $AP_{50}$ & $AP_{75}$ & $AR$\\
    \hline
    w/ & 88.1 & 98.9 & 97.3 & 92.9\\
    w/o & 86.2 & 96.4 & 95.6 & 91.9 \\
    \bottomrule[0.1em]
\end{tabular}
\label{tab: ablation1 position embedding}
\end{table}

\begin{table}[t]
 \centering
 \caption{Ablation studies on the visual encoder. In each pretrained model series, we select ViT-L for experiments to exclude the influence of vision encoder size.}
 \begin{tabular}{l|cccc}
    \toprule[0.2em]
     Method & $mAP$ & $AP_{50}$ & $AP_{75}$ & $AR$\\
	\hline
    RADIO & 88.1 & 98.9 & 97.3 & 92.9\\
    CLIP & 77.1 & 93.1 & 88.1 & 81.4 \\
    Dinov2 & 88.2 & 98.8 & 87.4 & 92.8 \\
    \bottomrule[0.1em]
\end{tabular}
\label{tab: ablation2 visual encoder}
\end{table}

\begin{table}[t]
 \centering
 \caption{Ablation studies on the effectiveness of various training stages.}
  \resizebox{0.5\textwidth}{!}{
 \begin{tabular}{ccc|cccc}
    \toprule[0.2em]
     Pretraining & SFT & DPO & $mAP$ & $AP_{50}$ & $AP_{75}$ & $AR$\\
    \hline
     &$\checkmark$& & 83.6 & 95.2 & 93.8 & 90.2 \\
    $\checkmark$& $\checkmark$ & & 86.3 & 97.1 & 95.9 & 91.8 \\
    $\checkmark$& $\checkmark$ & $\checkmark$ & 88.1 & 98.9 & 97.3 & 92.9\\
    \bottomrule[0.1em]
\end{tabular}
}
\label{tab: Ablation3: Tuning Strategy}
\end{table}

\begin{table}[t]
 \centering
 \caption{Ablation studies on the model size of the large language model.}
 \begin{tabular}{l|ccc}
    \toprule[0.2em]
     LLM & WHU & CrowdAI & WHU-Mix\\
	\hline
    Qwen3-0.6B & 88.1 AP & 86.1 AP & 78.0 AP \\
    Qwen3-1.7B & 90.2 AP & 87.7 AP & 78.5 AP \\
    \bottomrule[0.1em]
\end{tabular}
\label{tab: Ablation4: Large Language Model}
\end{table}

\begin{table}[t]
 \centering
 \caption{Ablation studies on the coordinate representation format.}
 \begin{tabular}{l|cccc}
    \toprule[0.2em]
     Method & $mAP$ & $AP_{50}$ & $AP_{75}$ & $AR$\\
	\hline
    Single Special Token & 88.1 & 98.9 & 97.3 & 92.9\\
    Multiple Special Tokens & 86.4 & 97.5 & 96.5 & 92.4\\
    \bottomrule[0.1em]
\end{tabular}
\label{tab: Ablation5: Coordinate Formulation}
\end{table}

\begin{table}[t]
 \centering
 \caption{Ablation studies on the effectiveness of data scaling.}
  \resizebox{0.5\textwidth}{!}{
 \begin{tabular}{cccc|cccc}
    \toprule[0.2em]
     WHU & COCO & Mix & CrowdAI & $mAP$ & $AP_{50}$ & $AP_{75}$ & $AR$\\
    \hline
    \checkmark & & & & 85.9 & 96.3 & 97.0 & 91.8 \\
    \checkmark & \checkmark & & & 86.2 & 96.7 & 97.2 & 92.1 \\
    \checkmark & \checkmark & \checkmark & \checkmark & 88.1 & 98.9 & 97.3 & 92.9 \\
    \bottomrule[0.1em]
\end{tabular}
}
\label{tab: Ablation6: Jointly Co-training}
\end{table}

\section{Experiments}
\label{sec:exp}

\subsection{Implementation Details}

We adopt the Qwen3-0.6B and Qwen3-1.7B~\cite{yang2025qwen3} as the LLM for VectorLLM-0.6B and VectorLLM-1.7B, and select RADIO-L~\cite{ranzinger2024radio} as the vision encoder.

\noindent\textbf{Training settings.} We pre-train VectorLLM using the aforementioned training data and loss through a three-stage process. During training, for each building, we randomly scale the ground-truth bounding box by 1.1 to 1.5 times as the cropping range to obtain cropped single building images. In the pretraining stage, we adopt a learning rate of 2e-4, with batch size set to 256 and warm-up ratio set to 0.03. VectorLLM is jointly trained on WHU~\cite{ji2018fully}, WHU-Mix~\cite{buildmapper}, CrowdAI~\cite{crowdai} and COCO~\cite{coco} training datasets for 24 epochs. In the SFT stage, the learning rate is set to 4e-5, batch size and warm-up ratio are set the same as the pretraining stage, but we only train for 4 epochs. In the DPO stage, we only use WHU data due to its superior annotation quality. In the DPO stage, the learning rate is set to 5e-7, batch size to 32, warm-up ratio to 0.03, and beta in the DPO loss is set to 0.5. All training is optimized using the AdamW optimizer.

\noindent\textbf{Test settings.} During testing, we adopt two settings. First, we use ground-truth bounding boxes (oracle box setting) to generate all building contour detection results for each image. Second, we train an additional detector to generate all building detection results for each image. For all bounding boxes, we apply a 1.3$\times$ enlargement as the cropping range to obtain single building images with certain backgrounds. We adopt FCOS~\cite{fcos} with Swin-L~\cite{swin} backbone as the detector. To accelerate testing, we employ batch inference with a batch size of 96, stopping inference once all batch samples produce end tokens and discarding predictions after the first end token in each sample.

\subsection{Main Results}

\noindent\textbf{WHU dataset.} We evaluate VectorLLM's performance on the test set of the WHU dataset, with results shown in Tab.~\ref{tab:whu}. Under the oracle bbox setting, VectorLLM achieves 90.2 AP, 98.9 AP$_{50}$, and 98.3 AP$_{75}$. Using detector-predicted bounding boxes, VectorLLM achieves 79.4 AP, 91.6 AP$_{50}$ and 85.9 AP$_{75}$. VectorLLM surpasses the previous SOTA methods P2PFormer~\cite{zhang2024p2pformer} and Line2Poly~\cite{wei2024lines} by 6.7 AP and 5.6 AP respectively, demonstrating significant improvements in extraction quality.

\noindent\textbf{WHU-Mix Dataset.} We evaluate VectorLLM's performance on the test-1 set (in-domain) and test-2 set (out-of-domain) of the WHU-Mix dataset, with results shown in Tab.~\ref{tab:whu-mix}. Using oracle bounding boxes, VectorLLM achieves 78.5 AP, 96.7 AP$_{50}$, and 86.0 AP$_{75}$ on the test-1 set, and 64.6 AP, 96.6 AP$_{50}$, and 74.2 AP$_{75}$ on the test-2 set. When using bounding boxes output by the detector, VectorLLM achieves 68.4 AP and 56.4 AP on the test-1 set and test-2 set respectively, surpassing the previous SOTA P2PFormer~\cite{zhang2024p2pformer} by 7.1 AP and 3.2 AP.

\noindent\textbf{CrowdAI Dataset.} We evaluate VectorLLM's performance on the CrowdAI dataset, as shown in Tab.~\ref{tab:crowdai}. There are some very unfavorable factors: first, CrowdAI images have low resolution and are relatively blurry, making it difficult to accurately identify building contours. Second, the annotation quality of the CrowdAI dataset is poor, with inconsistent standards where some building contours are annotated on rooftops while others on building bases. Despite of these, VectorLLM still achieves 87.7 AP and 79.6 AP using oracle bounding boxes and detector-output bounding boxes respectively. VectorLLM surpasses P2PFormer \cite{zhang2024p2pformer} by 13.6 AP.

\begin{figure*}[ht]
  \centering
  \includegraphics[width=1.0\linewidth]{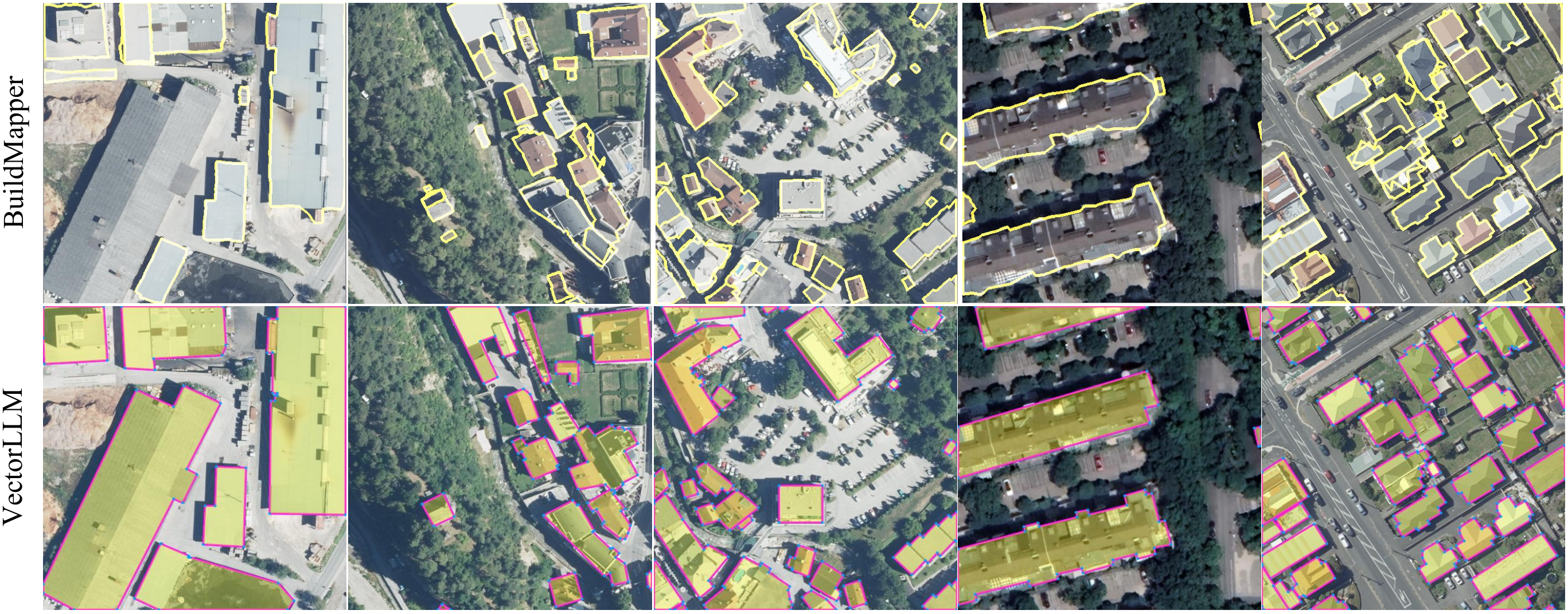}
  \includegraphics[width=1.0\linewidth]{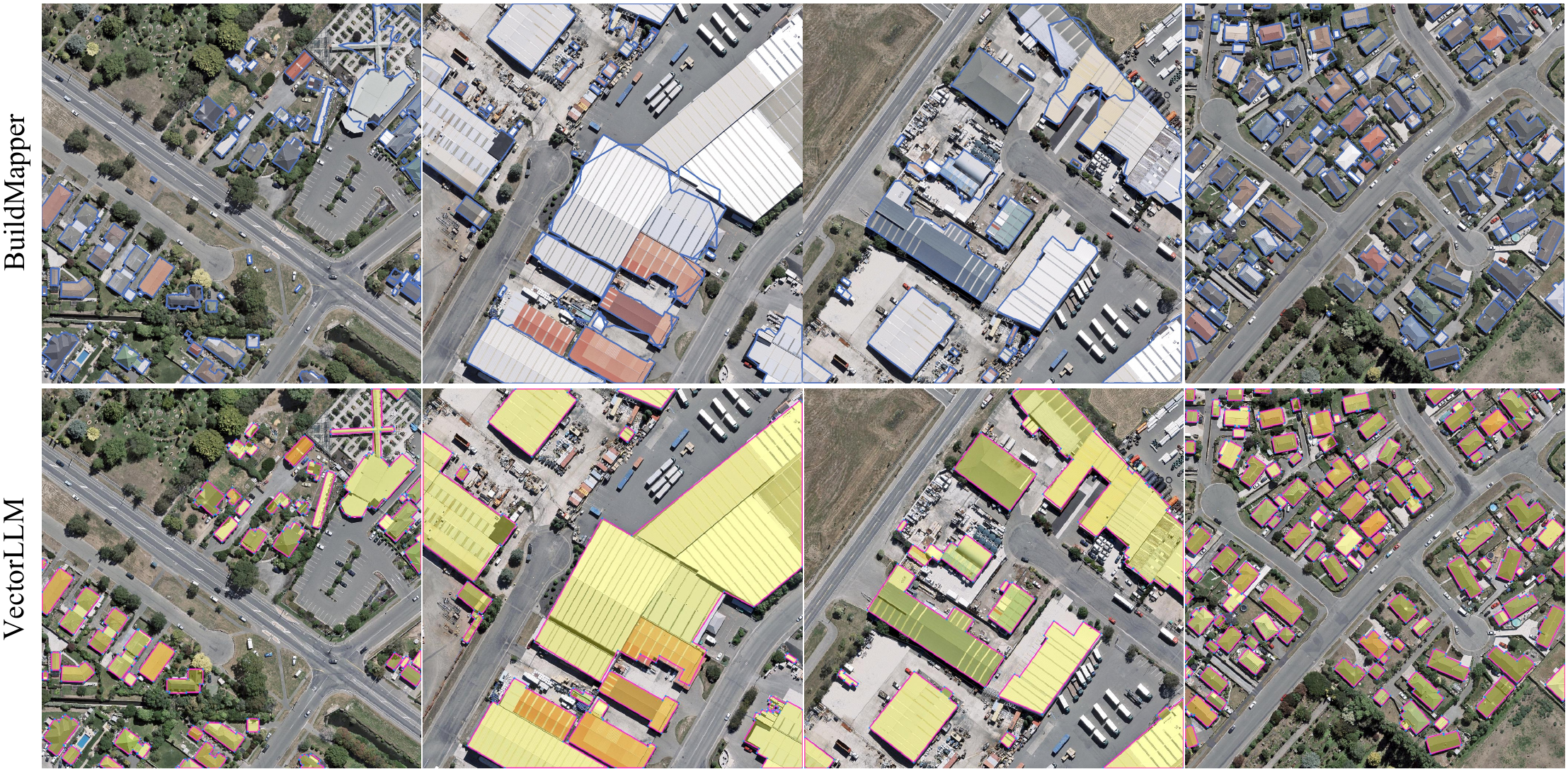}
  \caption{\textbf{The visualization results on WHU and WHU-Mix datasets.} VectorLLM demonstrates more stable and accurate vector extraction results compared to BuildMapper.}
  \label{fig:comparision_visualize_whu_whumix}
\end{figure*}

\subsection{Ablation Studies}
\label{sec:ablation}

We conduct ablation experiments on the test set of the WHU dataset, with all experiments using oracle bounding boxes to exclude additional bias introduced by detectors.

\noindent\textbf{Learnable Positional Embedding.} Learnable positional embedding can effectively inject position information into vision features. As shown in Tab.~\ref{tab: ablation1 position embedding}, when removing the learnable positional embedding component, VectorLLM exhibits a 1.9 AP performance drop.

\noindent\textbf{Vision Encoder.} The vision encoder is an important component of VectorLLM, responsible for extracting semantically rich features from images to provide to the LLM. We test three vision foundation models, including CLIP~\cite{radford2021learning}, DinoV2~\cite{oquab2023dinov2}, and RADIO~\cite{ranzinger2024radio}, with results shown in Tab.~\ref{tab: ablation2 visual encoder}. When using CLIP as the vision encoder, VectorLLM only achieves 77.1 AP, which is due to CLIP's pretraining approach causing it to lose substantial low-level information, which is crucial for vector extraction tasks. When using RADIO and DinoV2 as vision encoders, VectorLLM achieves similar performance. However, considering potential future extensions, such as enabling VectorLLM with VQA capabilities, maintaining alignment between the vision encoder and text is necessary, therefore we select RADIO as the vision encoder.

\begin{figure*}[ht]
  \centering
  \includegraphics[width=0.8\linewidth]{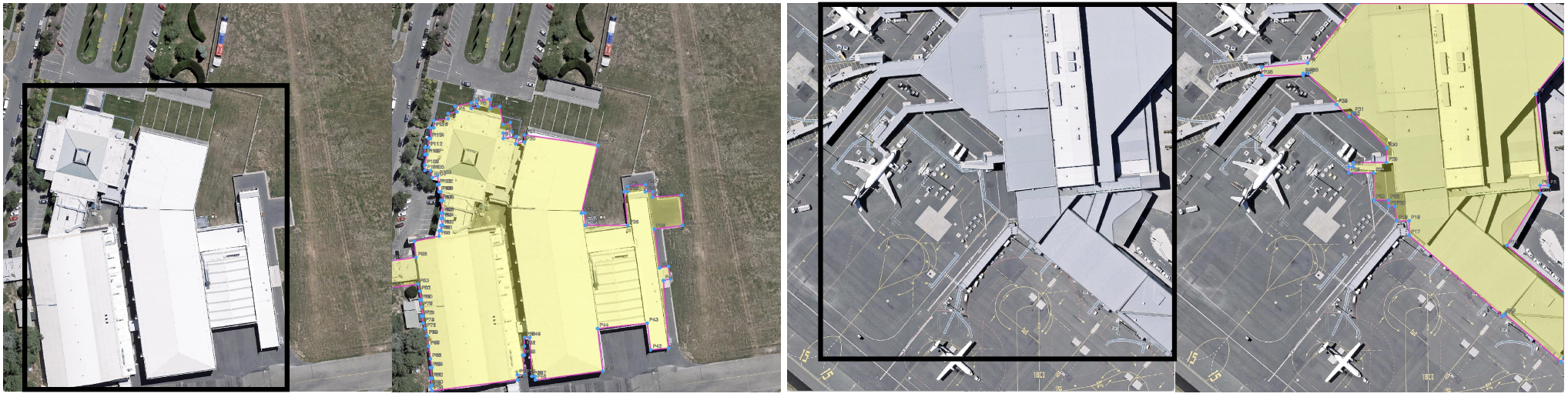}
  \caption{\textbf{Visualization results of failure cases.}}
  \label{fig:failure_cases}
\end{figure*}

\noindent\textbf{Tuning Strategy.} We explore the impact of various training stages on VectorLLM, with results shown in Tab.~\ref{tab: Ablation3: Tuning Strategy}. When removing the pretraining stage, VectorLLM exhibits significant performance degradation and training instability. This is because without aligning vision features to the LLM feature space, training leads to corruption of knowledge in both the vision encoder and LLM. The use of DPO brings performance improvements of 1.8 AP and 1.4 AP$_{75}$. We find that DPO significantly improves corner cases such as ignoring small edge lines and duplicate nodes.

\noindent\textbf{Large Language Model.} We try using different sizes of large language models to construct VectorLLM, with results shown in Tab~\ref{tab: Ablation4: Large Language Model}. When using larger-sized large language models, the model's performance improves, proving that model size scaling is still applicable in remote sensing vector extraction tasks. However, limited by available resources, we do not attempt larger sizes.

\noindent\textbf{Coordinate Formulation.} We explore the impact of different coordinate representation methods on VectorLLM's performance, as shown in Tab.~\ref{tab: Ablation5: Coordinate Formulation}. We try using multiple special tokens to separately represent the ones, tens, and hundreds digits of a coordinate, such as ``[hundreds-1][tens-2][ones-4]" to represent ``100+20+4=124". Although this representation method can improve the reuse rate of special tokens, it not only degrades VectorLLM's performance but also brings longer sequence lengths leading to increased computational cost. Therefore, we adopt the simplest representation method by assigning one special token for each value.

\noindent\textbf{Jointly Co-training.} We find that VectorLLM exhibits good data scaling properties. As shown in Tab.~\ref{tab: Ablation6: Jointly Co-training}, when using more data, VectorLLM's performance continues to improve, even though there are significant differences in domain, annotation standards, and image styles among these datasets.

\subsection{Visualization and Zero-shot Performance}

We conduct a qualitative analysis of VectorLLM. First, we compare the prediction results of BuildMapper and VectorLLM on WHU and WHU-Mix datasets, as shown in Fig.~\ref{fig:comparision_visualize_whu_whumix}. VectorLLM demonstrates more accurate vertex precision and significantly outperforms BuildMapper in terms of stability and accuracy for complex buildings. Then, we compare VectorLLM, P2PFormer, and PolyWorld on the CrowdAI dataset, as shown in Fig.~\ref{fig:comparision_visualize_crowdai}. VectorLLM exhibits more stable topological relationship prediction results, despite the lower image quality of CrowdAI greatly limiting fine-grained segmentation.

Furthermore, VectorLLM surprisingly demonstrates strong zero-shot prediction results on geographical features that never appear in the training set, as shown in Fig.~\ref{fig:other_rs_visualize}. VectorLLM also exhibits strong performance on general objects, even for multi-granularity data not included in the training set, as shown in Fig.~\ref{fig:general_obj_visualize}. Specifically, in the second image, VectorLLM represents the ground with sparse vertices due to its simple shape, but uses denser vertices for the motorcycle to depict its intricate structure, showcasing its dynamic handling ability of shape complexity.

\subsection{Failed cases}
Although VectorLLM demonstrates strong potential in vector extraction tasks, it still exhibits some limitations. As shown on the left side of Fig.~\ref{fig:failure_cases}, for large buildings, VectorLLM has significant edge localization bias, which is due to the adoption of limited resolution (128×128), causing detail loss of large buildings when compressed. Second, as shown on the right side of Fig.~\ref{fig:failure_cases}, for buildings with extremely complex structures, VectorLLM misses a large amount of detail. Additionally, VectorLLM cannot handle hollow buildings (which often require multiple polygons to represent jointly), nor can it process objects that are segmented into multiple parts.

%% file: latex/conclusion.tex
\section{Discussion}

\subsection{LLMs vs. conventional deep methods}
As shown in Tab.~\ref{tab:whu}, \ref{tab:whu-mix}, and \ref{tab:crowdai}, VectorLLM significantly outperforms SOTA methods by 5.6 AP, 7.1 AP, and 13.6 AP on the WHU, WHU-Mix and CrowdAI datasets, respectively. Taking the WHU dataset for example, the performance gaps among existing SOTA algorithms (e.g., BuildMapper, P2PFormer, Line2Poly) are merely ~1 mAP. A similar pattern is observed on WHU-Mix and CrowdAI datasets, leading to a key conclusion: LLM-based methods represent a qualitative leap over conventional deep learning approaches. This breakthrough parallels the seminal impact of AlexNet~\cite{krizhevsky2017imagenet} in early deep learning history, which shattered the performance ceiling of conventional methods. As the first work of its kind, our method establishes a new paradigm, with future refinements likely to yield even better results. These findings strongly suggest that LLM-based approaches may become the dominant framework for vector extraction tasks (e.g., building footprint delineation) in the foreseeable future.
\subsection{Our LLM for vectorization vs. current LLMs for remote sensing}
With the development of LLMs (or MLLM) in the field of computer vision, introducing LLMs to the remote sensing domain has become a natural trend. Image-text dialogue is the simplest case of multimodal applications, which is why current remote sensing LLMs begin with image-text dialogue for remote sensing images, such as~\cite{hu2025rsgpt,wang2024skyscript}. Other remote sensing LLMs offer object detection (grounding) capabilities, including~\cite{kuckreja2024geochat,zhang2024earthgpt}, following the object detection abilities already provided by LLMs in computer vision~\cite{zhang2025gpt4roi,rasheed2024glamm}. This paper takes a different starting point from these remote sensing LLMs. Our thinking begins with addressing a fundamental problem in high-resolution remote sensing image processing: the vector-form contour extraction of objects of interest. The most challenging task is land use and land cover (LULC) vector mapping. This requires an LLM to draw the vector boundaries of all ground objects using basic primitives like points and lines, just as a human would. This problem is overly difficult, so we chose to start our research with the simplest objects—buildings—and have achieved promising results. Our research approach can serve as a starting point for future LLM-based methods in LULC vector mapping.
\subsection{About generalization}
Researchers have gradually recognized that deep learning models, in fact, possess two levels of generalization capability. The first is cross-domain capability within the same task. For example, using a land cover segmentation model pretrained on one scenario to segment remote sensing images from unseen scenarios~\cite{luo2022cross}. The second is cross-task or cross-object capability, i.e., zero-shot ability. For instance, this paper demonstrates how a model pretrained on a building dataset can predict the polygonal contours of airplanes, lakes, backpacks, and other objects. Pursuing zero-shot capability is an inevitable trend in the development of LLMs. The original intention of LLMs is universality: we expect a general conversational LLM to answer questions about medicine, mathematics, daily life, and more. Naturally, we desire a remote sensing LLM for vectorization to handle the prediction of almost all types of ground objects. Our experiments have demonstrated VectorLLM’s superior zero-shot adaptability in remote sensing and close-range unknown object contour extraction, suggesting that LLM-based approaches are the right path to simultaneously address out-of-domain and zero-shot challenges.
\subsection{Future Work}
First, we will extend VectorLLM to more geographic feature types suitable for vector representation, such as roads, water bodies, and building roof fine structures. Second, we plan to use VectorLLM directly for dense segmentation, thereby removing the dependence on additional detectors. Finally, we will incorporate more high-quality data for data scaling (such as SAM~\cite{sam} data), enabling VectorLLM to acquire more powerful and stable vector extraction capabilities.
\section{Conclusion}
\label{sec:conclusion}

In this paper, we propose VectorLLM, the first model to use large language models for remote sensing fine-grained vector-form object contour extraction. We explore in detail how to perform pretraining, supervised finetuning, and preference optimization on MLLMs in the field of remote sensing vector extraction. Taking regular building vector extraction as an example, VectorLLM is jointly trained on multiple datasets and achieves SOTA performance on WHU, WHU-Mix, and CrowdAI datasets with a single model and weights. Additionally, VectorLLM demonstrates strong generalization capabilities, showing powerful zero-shot performance on various features including roads, ships, aircraft, playgrounds, vegetation, and water bodies. We believe VectorLLM has the potential to unify remote sensing vector extraction tasks, and we open-source all codes and weights to promote community development.

\noindent\textbf{Declaration of Interest}

The authors declare that they have no conflicts of interest.

\noindent\textbf{Acknowledgement}

This work was supported by the National Natural Science Foundation of China (42171430).

%% file: main.bbl
\begin{thebibliography}{xx}

\bibitem[Acuna et al., 2018]{polygonrnn++}
Acuna, D., Ling, H., Kar, A., Fidler, S., 2018.
 Efficient interactive annotation of segmentation datasets with polygon-rnn++.
 \emph{Proceedings of the IEEE conference on Computer Vision and Pattern Recognition}, 859--868.

\bibitem[Bai et al., 2023a]{bai2023qwen}
Bai, J., Bai, S., Chu, Y., Cui, Z., Dang, K., Deng, X., Fan, Y., Ge, W., Han, Y., Huang, F. et~al., 2023a.
 Qwen technical report.
 {\em arXiv preprint arXiv:2309.16609}.

\bibitem[Bai et al., 2023b]{bai2023qwenvl}
Bai, J., Bai, S., Yang, S., Wang, S., Tan, S., Wang, P., Lin, J., Zhou, C., Zhou, J., 2023b.
 Qwen-vl: A frontier large vision-language model with versatile abilities.
 {\em arXiv preprint arXiv:2308.12966}.

\bibitem[Bischke et al., 2019]{r5}
Bischke, B., Helber, P., Folz, J., Borth, D., Dengel, A., 2019.
 Multi-task learning for segmentation of building footprints with deep neural networks.
 \emph{2019 IEEE International Conference on Image Processing (ICIP)}, IEEE, 1480--1484.

\bibitem[Bolya et al., 2019]{yolact}
Bolya, D., Zhou, C., Xiao, F., Lee, Y.~J., 2019.
 Yolact: Real-time instance segmentation.
 \emph{Proceedings of the IEEE/CVF International Conference on Computer Vision}, 9157--9166.

\bibitem[Cai et al., 2024]{cai2024internlm2}
Cai, Z., Cao, M., Chen, H., Chen, K., Chen, K., Chen, X., Chen, X., Chen, Z., Chen, Z., Chu, P. et~al., 2024.
 Internlm2 technical report.
 {\em arXiv preprint arXiv:2403.17297}.

\bibitem[Castrejon et al., 2017]{polygonrnn}
Castrejon, L., Kundu, K., Urtasun, R., Fidler, S., 2017.
 Annotating object instances with a polygon-rnn.
 \emph{Proceedings of the IEEE conference on computer vision and pattern recognition}, 5230--5238.

\bibitem[Chen et al., 2022]{r6}
Chen, J., Jiang, Y., Luo, L., Gong, W., 2022.
 ASF-Net: Adaptive Screening Feature Network for Building Footprint Extraction From Remote-Sensing Images.
 {\em IEEE Transactions on Geoscience and Remote Sensing}, 60, 1--13.

\bibitem[Chen et al., 2024a]{chen2024rsprompter}
Chen, K., Liu, C., Chen, H., Zhang, H., Li, W., Zou, Z., Shi, Z., 2024a.
 RSPrompter: Learning to prompt for remote sensing instance segmentation based on visual foundation model.
 {\em IEEE Transactions on Geoscience and Remote Sensing}.

\bibitem[Chen et al., 2020]{r20}
Chen, Q., Wang, L., Waslander, S.~L., Liu, X., 2020.
 An end-to-end shape modeling framework for vectorized building outline generation from aerial images.
 {\em ISPRS Journal of Photogrammetry and Remote Sensing}, 170, 114--126.

\bibitem[Chen et al., 2021]{r7}
Chen, S., Shi, W., Zhou, M., Zhang, M., Xuan, Z., 2021.
 CGSANet: A Contour-Guided and Local Structure-Aware Encoder--Decoder Network for Accurate Building Extraction From Very High-Resolution Remote Sensing Imagery.
 {\em IEEE Journal of Selected Topics in Applied Earth Observations and Remote Sensing}, 15, 1526--1542.

\bibitem[Chen et al., 2024b]{chen2024expanding}
Chen, Z., Wang, W., Cao, Y., Liu, Y., Gao, Z., Cui, E., Zhu, J., Ye, S., Tian, H., Liu, Z. et~al., 2024b.
 Expanding Performance Boundaries of Open-Source Multimodal Models with Model, Data, and Test-Time Scaling.
 {\em arXiv preprint arXiv:2412.05271}.

\bibitem[Chen et al., 2024c]{chen2024far}
Chen, Z., Wang, W., Tian, H., Ye, S., Gao, Z., Cui, E., Tong, W., Hu, K., Luo, J., Ma, Z. et~al., 2024c.
 How Far Are We to GPT-4V? Closing the Gap to Commercial Multimodal Models with Open-Source Suites.
 {\em arXiv preprint arXiv:2404.16821}.

\bibitem[Chen et al., 2024d]{chen2024internvl}
Chen, Z., Wu, J., Wang, W., Su, W., Chen, G., Xing, S., Zhong, M., Zhang, Q., Zhu, X., Lu, L. et~al., 2024d.
 Internvl: Scaling up vision foundation models and aligning for generic visual-linguistic tasks.
 \emph{Proceedings of the IEEE/CVF Conference on Computer Vision and Pattern Recognition}, 24185--24198.

\bibitem[Cheng et al., 2022]{mask2former}
Cheng, B., Misra, I., Schwing, A.~G., Kirillov, A., Girdhar, R., 2022.
 Masked-attention mask transformer for universal image segmentation.
 \emph{Proceedings of the IEEE/CVF Conference on Computer Vision and Pattern Recognition}, 1290--1299.

\bibitem[Chiang et al., 2023]{vicuna2023}
Chiang, W.-L., Li, Z., Lin, Z., Sheng, Y., Wu, Z., Zhang, H., Zheng, L., Zhuang, S., Zhuang, Y., Gonzalez, J.~E., Stoica, I., Xing, E.~P., 2023.
 Vicuna: An open-source chatbot impressing gpt-4 with 90\%* chatgpt quality.

\bibitem[Dosovitskiy et al., 2020]{dosovitskiy2020image}
Dosovitskiy, A., Beyer, L., Kolesnikov, A., Weissenborn, D., Zhai, X., Unterthiner, T., Dehghani, M., Minderer, M., Heigold, G., Gelly, S. et~al., 2020.
 An image is worth 16x16 words: Transformers for image recognition at scale.
 {\em arXiv preprint arXiv:2010.11929}.

\bibitem[Dosovitskiy et al., 2021]{dosovitskiy2020vit}
Dosovitskiy, A., Beyer, L., Kolesnikov, A., Weissenborn, D., Zhai, X., Unterthiner, T., Dehghani, M., Minderer, M., Heigold, G., Gelly, S., Uszkoreit, J., Houlsby, N., 2021.
 An Image is Worth 16x16 Words: Transformers for Image Recognition at Scale.
 {\em ICLR}.

\bibitem[Dubey et al., 2024]{dubey2024llama}
Dubey, A., Jauhri, A., Pandey, A., Kadian, A., Al-Dahle, A., Letman, A., Mathur, A., Schelten, A., Yang, A., Fan, A. et~al., 2024.
 The llama 3 herd of models.
 {\em arXiv preprint arXiv:2407.21783}.

\bibitem[Elgendy et al., 2024]{elgendy2024geollava}
Elgendy, H., Sharshar, A., Aboeitta, A., Ashraf, Y., Guizani, M., 2024.
 Geollava: Efficient fine-tuned vision-language models for temporal change detection in remote sensing.
 {\em arXiv preprint arXiv:2410.19552}.

\bibitem[Fang et al., 2021]{queryinst}
Fang, Y., Yang, S., Wang, X., Li, Y., Fang, C., Shan, Y., Feng, B., Liu, W., 2021.
 Instances as queries.
 \emph{Proceedings of the IEEE/CVF International Conference on Computer Vision}, 6910--6919.

\bibitem[Floridi and Chiriatti, 2020]{floridi2020gpt}
Floridi, L., Chiriatti, M., 2020.
 GPT-3: Its nature, scope, limits, and consequences.
 {\em Minds and Machines}, 30, 681--694.

\bibitem[Girard et al., 2021]{ffl}
Girard, N., Smirnov, D., Solomon, J., Tarabalka, Y., 2021.
 Polygonal building extraction by frame field learning.
 \emph{Proceedings of the IEEE/CVF Conference on Computer Vision and Pattern Recognition}, 5891--5900.

\bibitem[Guo et al., 2024]{guo2024remote}
Guo, H., Su, X., Wu, C., Du, B., Zhang, L., Li, D., 2024.
 Remote sensing chatgpt: Solving remote sensing tasks with chatgpt and visual models.
 {\em arXiv preprint arXiv:2401.09083}.

\bibitem[He et al., 2024]{he2024highernet}
He, H., Ma, L., Li, J., 2024.
 HigherNet-DST: Higher resolution network with dynamic scale training for rooftop delineation.
 {\em IEEE Transactions on Geoscience and Remote Sensing}.

\bibitem[He et al., 2017]{maskrcnn}
He, K., Gkioxari, G., Doll{\'a}r, P., Girshick, R., 2017.
 Mask r-cnn.
 \emph{Proceedings of the IEEE international conference on computer vision}, 2961--2969.

\bibitem[Hu et al., 2025]{hu2025rsgpt}
Hu, Y., Yuan, J., Wen, C., Lu, X., Liu, Y., Li, X., 2025.
 Rsgpt: A remote sensing vision language model and benchmark.
 {\em ISPRS Journal of Photogrammetry and Remote Sensing}, 224, 272--286.

\bibitem[Huang et al., 2021a]{r25}
Huang, W., Liu, Z., Tang, H., Ge, J., 2021a.
 Sequentially delineation of rooftops with holes from VHR aerial images using a convolutional recurrent neural network.
 {\em Remote Sensing}, 13(21), 4271.

\bibitem[Huang et al., 2021b]{r23}
Huang, W., Tang, H., Xu, P., 2021b.
 OEC-RNN: Object-oriented delineation of rooftops with edges and corners using the recurrent neural network from the aerial images.
 {\em IEEE Transactions on Geoscience and Remote Sensing}, 60, 1--12.

\bibitem[Irvin et al., 2024]{irvin2024teochat}
Irvin, J.~A., Liu, E.~R., Chen, J.~C., Dormoy, I., Kim, J., Khanna, S., Zheng, Z., Ermon, S., 2024.
 Teochat: A large vision-language assistant for temporal earth observation data.
 {\em arXiv preprint arXiv:2410.06234}.

\bibitem[Ji et al., 2018]{ji2018fully}
Ji, S., Wei, S., Lu, M., 2018.
 Fully convolutional networks for multisource building extraction from an open aerial and satellite imagery data set.
 {\em IEEE Transactions on geoscience and remote sensing}, 57(1), 574--586.

\bibitem[Ji et al., 2019]{r3}
Ji, S., Wei, S., Lu, M., 2019.
 A scale robust convolutional neural network for automatic building extraction from aerial and satellite imagery.
 {\em International journal of remote sensing}, 40(9), 3308--3322.

\bibitem[Kirillov et al., 2023a]{sam}
Kirillov, A., Mintun, E., Ravi, N., Mao, H., Rolland, C., Gustafson, L., Xiao, T., Whitehead, S., Berg, A.~C., Lo, W.-Y. et~al., 2023a.
 Segment anything.
 {\em arXiv preprint arXiv:2304.02643}.

\bibitem[Kirillov et al., 2023b]{kirillov2023segment}
Kirillov, A., Mintun, E., Ravi, N., Mao, H., Rolland, C., Gustafson, L., Xiao, T., Whitehead, S., Berg, A.~C., Lo, W.-Y. et~al., 2023b.
 Segment anything.
 \emph{Proceedings of the IEEE/CVF International Conference on Computer Vision}, 4015--4026.

\bibitem[Krizhevsky et al., 2017]{krizhevsky2017imagenet}
Krizhevsky, A., Sutskever, I., Hinton, G.~E., 2017.
 ImageNet classification with deep convolutional neural networks.
 {\em Communications of the ACM}, 60(6), 84--90.

\bibitem[Kuckreja et al., 2024]{kuckreja2024geochat}
Kuckreja, K., Danish, M.~S., Naseer, M., Das, A., Khan, S., Khan, F.~S., 2024.
 Geochat: Grounded large vision-language model for remote sensing.
 \emph{Proceedings of the IEEE/CVF Conference on Computer Vision and Pattern Recognition}, 27831--27840.

\bibitem[Lai et al., 2024]{lai2024lisa}
Lai, X., Tian, Z., Chen, Y., Li, Y., Yuan, Y., Liu, S., Jia, J., 2024.
 Lisa: Reasoning segmentation via large language model.
 \emph{Proceedings of the IEEE/CVF Conference on Computer Vision and Pattern Recognition}, 9579--9589.

\bibitem[Li et al., 2024]{li2024vrsbench}
Li, X., Ding, J., Elhoseiny, M., 2024.
 VRSBench: A Versatile Vision-Language Benchmark Dataset for Remote Sensing Image Understanding.
 {\em arXiv preprint arXiv:2406.12384}.

\bibitem[Li et al., 2018]{polymapper}
Li, Z., Wegner, J.~D., Lucchi, A., 2018.
 Polymapper: Extracting city maps using polygons.
 {\em arXiv preprint arXiv:1812.01497}, 2.

\bibitem[Li et al., 2019]{r27}
Li, Z., Wegner, J.~D., Lucchi, A., 2019.
 Topological map extraction from overhead images.
 \emph{Proceedings of the IEEE/CVF International Conference on Computer Vision}, 1715--1724.

\bibitem[Liang et al., 2020]{polytransform}
Liang, J., Homayounfar, N., Ma, W.-C., Xiong, Y., Hu, R., Urtasun, R., 2020.
 Polytransform: Deep polygon transformer for instance segmentation.
 \emph{Proceedings of the IEEE/CVF Conference on Computer Vision and Pattern Recognition}, 9131--9140.

\bibitem[Lin et al., 2014]{coco}
Lin, T.-Y., Maire, M., Belongie, S., Hays, J., Perona, P., Ramanan, D., Doll{\'a}r, P., Zitnick, C.~L., 2014.
 Microsoft coco: Common objects in context.
 \emph{European conference on computer vision}, Springer, 740--755.

\bibitem[Lin et al., 2024]{lin2024drawandunderstand}
Lin, W., Wei, X., An, R., Gao, P., Zou, B., Luo, Y., Huang, S., Zhang, S., Li, H., 2024.
 Draw-and-understand: Leveraging visual prompts to enable mllms to comprehend what you want.

\bibitem[Liu et al., 2024]{liu2024change}
Liu, C., Chen, K., Zhang, H., Qi, Z., Zou, Z., Shi, Z., 2024.
 Change-agent: Towards interactive comprehensive remote sensing change interpretation and analysis.
 {\em IEEE Transactions on Geoscience and Remote Sensing}.

\bibitem[Liu et al., 2023a]{liu2023improvedllava}
Liu, H., Li, C., Li, Y., Lee, Y.~J., 2023a.
 Improved baselines with visual instruction tuning.

\bibitem[Liu et al., 2023b]{liu2023llava}
Liu, H., Li, C., Wu, Q., Lee, Y.~J., 2023b.
 Visual instruction tuning.

\bibitem[Liu et al., 2018]{panet}
Liu, S., Qi, L., Qin, H., Shi, J., Jia, J., 2018.
 Path aggregation network for instance segmentation.
 \emph{Proceedings of the IEEE conference on computer vision and pattern recognition}, 8759--8768.

\bibitem[Liu et al., 2021]{swin}
Liu, Z., Lin, Y., Cao, Y., Hu, H., Wei, Y., Zhang, Z., Lin, S., Guo, B., 2021.
 Swin transformer: Hierarchical vision transformer using shifted windows.
 \emph{Proceedings of the IEEE/CVF international conference on computer vision}, 10012--10022.

\bibitem[Liu et al., 2022]{r26}
Liu, Z., Tang, H., Huang, W., 2022.
 Building Outline Delineation From VHR Remote Sensing Images Using the Convolutional Recurrent Neural Network Embedded With Line Segment Information.
 {\em IEEE Transactions on Geoscience and Remote Sensing}, 60, 1--13.

\bibitem[Luo et al., 2024a]{luo2024skysensegpt}
Luo, J., Pang, Z., Zhang, Y., Wang, T., Wang, L., Dang, B., Lao, J., Wang, J., Chen, J., Tan, Y. et~al., 2024a.
 Skysensegpt: A fine-grained instruction tuning dataset and model for remote sensing vision-language understanding.
 {\em arXiv preprint arXiv:2406.10100}.

\bibitem[Luo and Ji, 2022]{luo2022cross}
Luo, M., Ji, S., 2022.
 Cross-spatiotemporal land-cover classification from VHR remote sensing images with deep learning based domain adaptation.
 {\em ISPRS Journal of Photogrammetry and Remote Sensing}, 191, 105--128.

\bibitem[Luo et al., 2024b]{luo2024sam}
Luo, M., Zhang, T., Wei, S., Ji, S., 2024b.
 SAM-RSIS: Progressively adapting SAM with box prompting to remote sensing image instance segmentation.
 {\em IEEE Transactions on Geoscience and Remote Sensing}.

\bibitem[Mohanty, 2019 (accessed November 10, 2019)]{crowdai}
Mohanty, S.~P., 2019 (accessed November 10, 2019).
 CrowdAI mapping challenge 2018 dataset.
 \url{https://www.crowdai.org/challenges/mapping-challenge}.

\bibitem[Muhtar et al., 2024]{muhtar2024lhrs}
Muhtar, D., Li, Z., Gu, F., Zhang, X., Xiao, P., 2024.
 Lhrs-bot: Empowering remote sensing with vgi-enhanced large multimodal language model.
 {\em arXiv preprint arXiv:2402.02544}.

\bibitem[Oquab et al., 2023]{oquab2023dinov2}
Oquab, M., Darcet, T., Moutakanni, T., Vo, H., Szafraniec, M., Khalidov, V., Fernandez, P., Haziza, D., Massa, F., El-Nouby, A. et~al., 2023.
 Dinov2: Learning robust visual features without supervision.
 {\em arXiv preprint arXiv:2304.07193}.

\bibitem[Pang et al., 2024]{pang2024h2rsvlm}
Pang, C., Wu, J., Li, J., Liu, Y., Sun, J., Li, W., Weng, X., Wang, S., Feng, L., Xia, G.-S. et~al., 2024.
 H2RSVLM: Towards Helpful and Honest Remote Sensing Large Vision Language Model.
 {\em arXiv preprint arXiv:2403.20213}.

\bibitem[Peng et al., 2020]{deepsnake}
Peng, S., Jiang, W., Pi, H., Li, X., Bao, H., Zhou, X., 2020.
 Deep snake for real-time instance segmentation.
 \emph{Proceedings of the IEEE/CVF Conference on Computer Vision and Pattern Recognition}, 8533--8542.

\bibitem[Radford et al., 2021]{radford2021learning}
Radford, A., Kim, J.~W., Hallacy, C., Ramesh, A., Goh, G., Agarwal, S., Sastry, G., Askell, A., Mishkin, P., Clark, J. et~al., 2021.
 Learning transferable visual models from natural language supervision.
 \emph{International conference on machine learning}, PMLR, 8748--8763.

\bibitem[Rafailov et al., 2024]{rafailov2024direct}
Rafailov, R., Sharma, A., Mitchell, E., Manning, C.~D., Ermon, S., Finn, C., 2024.
 Direct preference optimization: Your language model is secretly a reward model.
 {\em Advances in Neural Information Processing Systems}, 36.

\bibitem[Ranzinger et al., 2024]{ranzinger2024radio}
Ranzinger, M., Heinrich, G., Kautz, J., Molchanov, P., 2024.
 Am-radio: Agglomerative vision foundation model reduce all domains into one.
 \emph{Proceedings of the IEEE/CVF Conference on Computer Vision and Pattern Recognition}, 12490--12500.

\bibitem[Rasheed et al., 2024]{rasheed2024glamm}
Rasheed, H., Maaz, M., Shaji, S., Shaker, A., Khan, S., Cholakkal, H., Anwer, R.~M., Xing, E., Yang, M.-H., Khan, F.~S., 2024.
 Glamm: Pixel grounding large multimodal model.
 \emph{Proceedings of the IEEE/CVF Conference on Computer Vision and Pattern Recognition}, 13009--13018.

\bibitem[Ren et al., 2024]{ren2024pixellm}
Ren, Z., Huang, Z., Wei, Y., Zhao, Y., Fu, D., Feng, J., Jin, X., 2024.
 Pixellm: Pixel reasoning with large multimodal model.
 \emph{Proceedings of the IEEE/CVF Conference on Computer Vision and Pattern Recognition}, 26374--26383.

\bibitem[Singh et al., 2024]{singh2024geollm}
Singh, S., Fore, M., Stamoulis, D., 2024.
 Geollm-engine: A realistic environment for building geospatial copilots.
 \emph{Proceedings of the IEEE/CVF Conference on Computer Vision and Pattern Recognition}, 585--594.

\bibitem[Team, 2023]{team2023internlm}
Team, I., 2023.
 Internlm: A multilingual language model with progressively enhanced capabilities.

\bibitem[Tian et al., 2019]{fcos}
Tian, Z., Shen, C., Chen, H., He, T., 2019.
 Fcos: Fully convolutional one-stage object detection.
 \emph{Proceedings of the IEEE/CVF international conference on computer vision}, 9627--9636.

\bibitem[Touvron et al., 2023]{touvron2023llama}
Touvron, H., Martin, L., Stone, K., Albert, P., Almahairi, A., Babaei, Y., Bashlykov, N., Batra, S., Bhargava, P., Bhosale, S. et~al., 2023.
 Llama 2: Open foundation and fine-tuned chat models.
 {\em arXiv preprint arXiv:2307.09288}.

\bibitem[Wang et al., 2023]{wang2023image}
Wang, J., Ji, S., Zhang, T., 2023.
 From image transfer to object transfer: Cross-domain instance segmentation based on center point feature alignment.
 {\em IEEE Transactions on Geoscience and Remote Sensing}, 61, 1--11.

\bibitem[Wang et al., 2024a]{Qwen2VL}
Wang, P., Bai, S., Tan, S., Wang, S., Fan, Z., Bai, J., Chen, K., Liu, X., Wang, J., Ge, W., Fan, Y., Dang, K., Du, M., Ren, X., Men, R., Liu, D., Zhou, C., Zhou, J., Lin, J., 2024a.
 Qwen2-VL: Enhancing Vision-Language Model's Perception of the World at Any Resolution.
 {\em arXiv preprint arXiv:2409.12191}.

\bibitem[Wang et al., 2024b]{wang2024ringmogpt}
Wang, P., Hu, H., Tong, B., Zhang, Z., Yao, F., Feng, Y., Zhu, Z., Chang, H., Diao, W., Ye, Q. et~al., 2024b.
 RingMoGPT: A Unified Remote Sensing Foundation Model for Vision, Language, and grounded tasks.
 {\em IEEE Transactions on Geoscience and Remote Sensing}.

\bibitem[Wang et al., 2020]{solo}
Wang, X., Kong, T., Shen, C., Jiang, Y., Li, L., 2020.
 Solo: Segmenting objects by locations.
 \emph{European Conference on Computer Vision}, Springer, 649--665.

\bibitem[Wang et al., 2024c]{wang2024skyscript}
Wang, Z., Prabha, R., Huang, T., Wu, J., Rajagopal, R., 2024c.
 Skyscript: A large and semantically diverse vision-language dataset for remote sensing.
 \emph{Proceedings of the AAAI Conference on Artificial Intelligence}, ~38number~6, 5805--5813.

\bibitem[Wei and Ji, 2021]{r34}
Wei, S., Ji, S., 2021.
 Graph convolutional networks for the automated production of building vector maps from aerial images.
 {\em IEEE Transactions on Geoscience and Remote Sensing}, 60, 1--11.

\bibitem[Wei et al., 2019]{mafcn}
Wei, S., Ji, S., Lu, M., 2019.
 Toward automatic building footprint delineation from aerial images using CNN and regularization.
 {\em IEEE Transactions on Geoscience and Remote Sensing}, 58(3), 2178--2189.

\bibitem[Wei et al., 2021]{clpcnn}
Wei, S., Zhang, T., Ji, S., 2021.
 A Concentric Loop Convolutional Neural Network for Manual Delineation-Level Building Boundary Segmentation From Remote-Sensing Images.
 {\em IEEE Transactions on Geoscience and Remote Sensing}, 60, 1--11.

\bibitem[Wei et al., 2023]{buildmapper}
Wei, S., Zhang, T., Ji, S., Luo, M., Gong, J., 2023.
 BuildMapper: A fully learnable framework for vectorized building contour extraction.
 {\em ISPRS Journal of Photogrammetry and Remote Sensing}, 197, 87--104.

\bibitem[Wei et al., 2024]{wei2024lines}
Wei, S., Zhang, T., Yu, D., Ji, S., Zhang, Y., Gong, J., 2024.
 From lines to Polygons: Polygonal building contour extraction from High-Resolution remote sensing imagery.
 {\em ISPRS Journal of Photogrammetry and Remote Sensing}, 209, 213--232.

\bibitem[Wu et al., 2022]{r8}
Wu, Y., Xu, L., Chen, Y., Wong, A., Clausi, D.~A., 2022.
 TAL: Topography-Aware Multi-Resolution Fusion Learning for Enhanced Building Footprint Extraction.
 {\em IEEE Geoscience and Remote Sensing Letters}, 19, 1--5.

\bibitem[Xia et al., 2024a]{xia2024vectorizing}
Xia, X., Zhang, T., Heitzler, M., Hurni, L., 2024a.
 Vectorizing historical maps with topological consistency: A hybrid approach using transformers and contour-based instance segmentation.
 {\em International Journal of Applied Earth Observation and Geoinformation}, 129, 103837.

\bibitem[Xia et al., 2024b]{xia2024video}
Xia, X., Zhang, T., Hurni, L., 2024b.
 Video instance segmentation is all you need for linking geographic entities from historical maps.
 \emph{IGARSS 2024-2024 IEEE International Geoscience and Remote Sensing Symposium}, IEEE, 8491--8494.

\bibitem[Xie et al., 2020]{polarmask}
Xie, E., Sun, P., Song, X., Wang, W., Liu, X., Liang, D., Shen, C., Luo, P., 2020.
 Polarmask: Single shot instance segmentation with polar representation.
 \emph{Proceedings of the IEEE/CVF conference on computer vision and pattern recognition}, 12193--12202.

\bibitem[Xu et al., 2023]{xu2023hisup}
Xu, B., Xu, J., Xue, N., Xia, G.-S., 2023.
 HiSup: Accurate polygonal mapping of buildings in satellite imagery with hierarchical supervision.
 {\em ISPRS Journal of Photogrammetry and Remote Sensing}, 198, 284--296.

\bibitem[Yang et al., 2025]{yang2025qwen3}
Yang, A., Li, A., Yang, B., Zhang, B., Hui, B., Zheng, B., Yu, B., Gao, C., Huang, C., Lv, C. et~al., 2025.
 Qwen3 technical report.
 {\em arXiv preprint arXiv:2505.09388}.

\bibitem[Yang et al., 2024]{yang2024qwen2}
Yang, A., Yang, B., Hui, B., Zheng, B., Yu, B., Zhou, C., Li, C., Li, C., Liu, D., Huang, F. et~al., 2024.
 Qwen2 technical report.
 {\em arXiv preprint arXiv:2407.10671}.

\bibitem[Yu et al., 2019]{r28}
Yu, Y., Si, X., Hu, C., Zhang, J., 2019.
 A review of recurrent neural networks: LSTM cells and network architectures.
 {\em Neural computation}, 31(7), 1235--1270.

\bibitem[Yuan et al., 2025]{yuan2025sa2va}
Yuan, H., Li, X., Zhang, T., Huang, Z., Xu, S., Ji, S., Tong, Y., Qi, L., Feng, J., Yang, M.-H., 2025.
 Sa2VA: Marrying SAM2 with LLaVA for Dense Grounded Understanding of Images and Videos.
 {\em arXiv preprint arXiv:2501.04001}.

\bibitem[Yuan, 2017]{r4}
Yuan, J., 2017.
 Learning building extraction in aerial scenes with convolutional networks.
 {\em IEEE transactions on pattern analysis and machine intelligence}, 40(11), 2793--2798.

\bibitem[Yuan et al., 2024]{yuan2024osprey}
Yuan, Y., Li, W., Liu, J., Tang, D., Luo, X., Qin, C., Zhang, L., Zhu, J., 2024.
 Osprey: Pixel understanding with visual instruction tuning.
 \emph{Proceedings of the IEEE/CVF Conference on Computer Vision and Pattern Recognition}, 28202--28211.

\bibitem[Zhan et al., 2024]{zhan2024skyeyegpt}
Zhan, Y., Xiong, Z., Yuan, Y., 2024.
 Skyeyegpt: Unifying remote sensing vision-language tasks via instruction tuning with large language model.
 {\em arXiv preprint arXiv:2401.09712}.

\bibitem[Zhang et al., 2024a]{zhang2024hit}
Zhang, M., Liu, Q., Wang, Y., 2024a.
 HiT: Building Mapping with Hierarchical Transformers.
 {\em IEEE Transactions on Geoscience and Remote Sensing}.

\bibitem[Zhang et al., 2025a]{zhang2025gpt4roi}
Zhang, S., Sun, P., Chen, S., Xiao, M., Shao, W., Zhang, W., Liu, Y., Chen, K., Luo, P., 2025a.
 Gpt4roi: Instruction tuning large language model on region-of-interest.
 \emph{European Conference on Computer Vision}, Springer, 52--70.

\bibitem[Zhang et al., 2024b]{zhang2024omg}
Zhang, T., Li, X., Fei, H., Yuan, H., Wu, S., Ji, S., Loy, C.~C., Yan, S., 2024b.
 Omg-llava: Bridging image-level, object-level, pixel-level reasoning and understanding.
 {\em arXiv preprint arXiv:2406.19389}.

\bibitem[Zhang et al., 2025b]{zhang2025pixel}
Zhang, T., Li, X., Huang, Z., Li, Y., Lei, W., Deng, X., Chen, S., Ji, S., Feng, J., 2025b.
 Pixel-sail: Single transformer for pixel-grounded understanding.
 {\em arXiv preprint arXiv:2504.10465}.

\bibitem[Zhang et al., 2022]{e2ec}
Zhang, T., Wei, S., Ji, S., 2022.
 E2ec: An end-to-end contour-based method for high-quality high-speed instance segmentation.
 \emph{Proceedings of the IEEE/CVF Conference on Computer Vision and Pattern Recognition}, 4443--4452.

\bibitem[Zhang et al., 2024c]{zhang2024p2pformer}
Zhang, T., Wei, S., Zhou, Y., Luo, M., Yu, W., Ji, S., 2024c.
 P2PFormer: A Primitive-to-polygon Method for Regular Building Contour Extraction from Remote Sensing Images.
 {\em IEEE Transactions on Geoscience and Remote Sensing}.

\bibitem[Zhang et al., 2024d]{zhang2024popeye}
Zhang, W., Cai, M., Zhang, T., Lei, G., Zhuang, Y., Mao, X., 2024d.
 Popeye: A Unified Visual-Language Model for Multi-Source Ship Detection from Remote Sensing Imagery.
 {\em arXiv preprint arXiv:2403.03790}.

\bibitem[Zhang et al., 2024e]{zhang2024earthgpt}
Zhang, W., Cai, M., Zhang, T., Zhuang, Y., Mao, X., 2024e.
 Earthgpt: A universal multi-modal large language model for multi-sensor image comprehension in remote sensing domain.
 {\em IEEE Transactions on Geoscience and Remote Sensing}.

\bibitem[Zhang et al., 2024f]{zhang2024rs5m}
Zhang, Z., Zhao, T., Guo, Y., Yin, J., 2024f.
 RS5M and GeoRSCLIP: A large scale vision-language dataset and a large vision-language model for remote sensing.
 {\em IEEE Transactions on Geoscience and Remote Sensing}.

\bibitem[Zhao et al., 2018]{r17}
Zhao, K., Kang, J., Jung, J., Sohn, G., 2018.
 Building extraction from satellite images using mask r-cnn with building boundary regularization.
 \emph{Proceedings of the IEEE conference on computer vision and pattern recognition workshops}, 247--251.

\bibitem[Zhao et al., 2020]{r16}
Zhao, W., Persello, C., Stein, A., 2020.
 Building instance segmentation and boundary regularization from high-resolution remote sensing images.
 \emph{IGARSS 2020-2020 IEEE International Geoscience and Remote Sensing Symposium}, IEEE, 3916--3919.

\bibitem[Zhao et al., 2021]{r24}
Zhao, W., Persello, C., Stein, A., 2021.
 Building outline delineation: From aerial images to polygons with an improved end-to-end learning framework.
 {\em ISPRS journal of photogrammetry and remote sensing}, 175, 119--131.

\bibitem[Zhou et al., 2024]{zhou2024geoground}
Zhou, Y., Lan, M., Li, X., Ke, Y., Jiang, X., Feng, L., Zhang, W., 2024.
 GeoGround: A Unified Large Vision-Language Model. for Remote Sensing Visual Grounding.
 {\em arXiv preprint arXiv:2411.11904}.

\bibitem[Zhu et al., 2025]{zhu2025internvl3}
Zhu, J., Wang, W., Chen, Z., Liu, Z., Ye, S., Gu, L., Duan, Y., Tian, H., Su, W., Shao, J. et~al., 2025.
 InternVL3: Exploring Advanced Training and Test-Time Recipes for Open-Source Multimodal Models.
 {\em arXiv preprint arXiv:2504.10479}.

\bibitem[Zhu et al., 2020]{r9}
Zhu, Q., Liao, C., Hu, H., Mei, X., Li, H., 2020.
 MAP-Net: Multiple attending path neural network for building footprint extraction from remote sensed imagery.
 {\em IEEE Transactions on Geoscience and Remote Sensing}, 59(7), 6169--6181.

\bibitem[Zorzi et al., 2022]{polyworld}
Zorzi, S., Bazrafkan, S., Habenschuss, S., Fraundorfer, F., 2022.
 Polyworld: Polygonal building extraction with graph neural networks in satellite images.
 \emph{Proceedings of the IEEE/CVF Conference on Computer Vision and Pattern Recognition}, 1848--1857.

\bibitem[Zorzi et al., 2021]{r18}
Zorzi, S., Bittner, K., Fraundorfer, F., 2021.
 Machine-learned regularization and polygonization of building segmentation masks.
 \emph{2020 25th International Conference on Pattern Recognition (ICPR)}, IEEE, 3098--3105.

\end{thebibliography}
